\documentclass{article}
\usepackage[utf8]{inputenc}

\usepackage{enumitem}
\usepackage{amsmath,amssymb,amsfonts}
\usepackage{graphicx}
\usepackage{textcomp}
\usepackage{xcolor}
\usepackage{bm}
\usepackage{tabularx}
\usepackage{booktabs}
\usepackage{multirow}
\usepackage{caption}
\usepackage{subcaption}
\usepackage{url}
\usepackage{algorithm}
\usepackage{algpseudocode}
\usepackage{tikz}
\usetikzlibrary{positioning, calc, fit, shapes, arrows, external}

\usepackage{authblk}

\usepackage [english]{babel}
\usepackage [autostyle, english = american]{csquotes}
\MakeOuterQuote{"}

\usepackage{geometry}
\usepackage{setspace}
\usepackage[
backend=biber,
style=apa
]{biblatex}
\title{Assurance Monitoring of Learning Enabled Cyber-Physical Systems Using Inductive Conformal Prediction based on Distance Learning}

\author[]{Dimitrios Boursinos}
\author[]{Xenofon Koutsoukos}
\affil[]{Department of Electrical Engineering and Computer Science \\ Institute for Software Integrated Systems \\ Vanderbilt University, Nashville TN, USA\\
\textit{\{dimitrios.boursinos, xenofon.koutsoukos\}@vanderbilt.edu}}

\date{}

\begin{document}

\maketitle

\section*{Abstract}
\label{sec:abstract}
\textit{Machine learning components such as deep neural networks are used extensively in Cyber-Physical Systems (CPS). However, such components may introduce new types of hazards that can have disastrous consequences and need to be addressed for engineering trustworthy systems. 
Although deep neural networks offer advanced capabilities, they must be complemented by engineering methods and practices that allow effective integration in CPS. 
In this paper, we proposed an approach for assurance monitoring of learning-enabled CPS based on the conformal prediction framework.
In order to allow real-time assurance monitoring, the approach employs distance learning to transform high-dimensional inputs into lower size embedding representations.
By leveraging conformal prediction, the approach provides well-calibrated confidence and ensures a bounded small error rate while limiting the number of inputs for which an accurate prediction cannot be made.
We demonstrate the approach using three data sets of mobile robot following a wall, speaker recognition, and traffic sign recognition. 
The experimental results demonstrate that the error rates are well-calibrated while the number of 
alarms is very small. Further, the method is computationally efficient and allows real-time assurance monitoring of CPS. 
}

\section*{Keywords}
cyber-physical systems; deep neural networks; assurance monitoring; conformal prediction; prediction confidence

\section{Introduction}
\label{sec:introduction}

% Motivation
Cyber-Physical systems (CPS) can benefit by incorporating machine learning components that can 
handle the uncertainty and variability of the real-world. Typical components such as deep neural networks (DNNs)
can be used for performing various tasks such as perception of the environment. 
In autonomous vehicles, for example,  perception components aim at making sense of the surroundings like recognizing correctly traffic signs.
However, such DNNs introduce new types of hazards that can have disastrous consequences and need to be addressed for engineering trustworthy systems. 
Although DNNs offer advanced capabilities, they must be complemented by engineering methods and practices that allow effective integration in CPS. 

% Hazards
 A DNN is designed using learning techniques that require
specification of the task, a measure for evaluating how well the task is performed, and experience 
which typically includes training and testing data. Using the DNN during system operation 
presents challenges that must be addressed using innovative engineering methods. 
Perception of the environment is a functionality that is difficult to specify, and typically, specifications
are based on examples.
DNNs exhibit some nonzero error rate, the true error rate is unknown, and only an estimate from a design-time 
statistical process is known.
Further, DNNs encode information in a complex manner and it is hard to reason about the encoding.
Non-transparency is an obstacle to monitoring because it is more difficult to have confidence that the 
model is operating as intended.

% Problem considered in this paper
Our objective in this paper is to complement the prediction of DNNs
 with a computation of confidence that can be used for decision making. We consider DNNs used for classification in CPS.
In addition to the class prediction, we compute set predictors with a given confidence using 
 the conformal prediction framework~\parencite{balasubramanian2014conformal}. 
 We focus on computationally efficient algorithms that can be used for real-time monitoring. 
 An efficient and robust approach must ensure a small and well-calibrated error rate while limiting 
 the number of alarms. This enables the design of monitors which can ensure a bounded small error 
rate while limiting the number of inputs for which an accurate prediction cannot be made.

% Conformal prediction
The proposed approach is based on conformal prediction (CP)~\parencite{Vovk:2005:ALR:1062391,balasubramanian2014conformal}.
CP aims at associating reliable measures of confidence with set predictions for problems that include classification and
regression. An important feature of the CP framework is the calibration of the obtained confidence
values in an online setting which is very promising for real-time monitoring in CPS applications.
These methods can be applied for a variety of machine learning algorithms that include DNNs. 
The main idea is to test if a new input example
conforms to the training data set by utilizing a \textit{nonconformity measure} (NCM)
which assigns a numerical score indicating how different the input example 
is from the training data set. 
The next step is to define a $p$-value as the fraction of
observations that have nonconformity (NC) scores greater than or equal to the NC scores of the training examples which is then used for estimating the confidence of the prediction for the test input. 
In order to use the approach online, inductive conformal prediction (ICP) has been developed for computational efficiency~\parencite{papadopoulos2007conformal, balasubramanian2014conformal}. 
In ICP, the training dataset is split into the proper training dataset that is used for learning and 
a calibration dataset that is used to compute the predictions for given confidence levels. Existing methods rely on NCMs computed using techniques such as $k$-Nearest Neighbors and Kernel Density Estimation 
and do not scale for high-dimensional inputs in CPS.

%TMCE paper
DNNs have the ability to compute layers of representations of the input data which can then be used to distinguish between available classes~\parencite{hinton2007learning,bengio2009learning}. In our previous work, we developed an approach for mapping high-dimensional inputs into lower-dimensional representations to make the application of ICP possible for assurance monitoring of CPS in real-time~\parencite{boursinos2020assurance}. 
The approach utilizes the vector of the neuron activations in the penultimate layer of the DNN for a particular input. This low-dimensional representation can be used to compute NC scores efficiently for high-dimensional inputs. In problems where the input data are high-dimensional, such as the classification of traffic sign images in autonomous vehicles, ICP based on these learned embedding representations produces confident predictions. Moreover the execution time and the required memory is significantly lower than using the original inputs and the approach can be used for real-time assurance monitoring of the DNN.
The use of low-dimensional learned embedding representations results in improved performance compared with ICP based on the original inputs. However, the  underlying DNN is still trained to perform classification and does not learn necessarily optimal representations for computing NC scores. 

% Express the novelty of the scientific problem and the approach with more emphasis in the Introduction
% This paper
The main challenge addressed in this paper is the efficient computation of embedding representations that allows assurance monitoring based on conformal prediction in real-time. The novelty of the approach lies on using distance metric learning to generate representations of the input data and use Euclidean distance as a measure of similarity. Unlike training a classifier where each training input is assigned a ground truth label and the objective is to minimize a loss function so that the prediction of the classifier will be the same as the label, in distance metric learning, the inputs are considered in pairs. The associated loss function is defined using pairwise constraints such that its minimization will make representations of inputs that belong to the same class be close to each other and representations of inputs belonging to different classes be far from each other. 
% The representation learning architectures we use are the siamese~\parencite{koch2015siamese} and triplet networks~\parencite{hoffer2014deep}. 
Preliminary results on using appropriate representations for a robotic navigation benchmark with low-dimensional inputs are presented in~\parencite{boursinos2020trusted}. 

% XK: Contributions: I think we need a list with 3 long paragraphs: (1) metric learning + ICP, (2) determining significance level and assurance monitoring algorithm with explanation of \Gamma, and (3) empirical evaluation.

% XK: We will need to extend this paragraph. We can make this contribution more technical by providing details:
% - move the sentence about the architectures from the paragraph above here
% - Describe the NC functions

The main contribution of the paper is the leverage of distance metric learning for assurance monitoring of learning-enabled CPS. The proposed approach based on ICP can be used in real time for high-dimensional data that are typically used in CPS. Different NC functions can be used in ICP to evaluate whether new unknown inputs are similar to the data that have been used for training a learning-enabled component such as a DNN. A NC function assigns a score to a labeled input reflecting how well it conforms to the training dataset. 
Because the choice of the NC function is very important, the proposed approach utilizes neural network architectures for distance metric learning based on siamese~\parencite{koch2015siamese} and triplet networks~\parencite{hoffer2014deep} to learn representations and define NC functions based on Euclidean distance. 
Specifically, the proposed functions compute the NC scores of a new labeled input using (1) the labels of its closest neighbors, (2) how far the closest neighbor of the same class is compared to any other neighbor, and (3) how far the label's centroid % XK: Is that correct, I did not know what 'it' is so I rephrased.
is compared to the centroids of the other labels. The main benefit of the approach is that by utilizing distance metric learning in ICP, we reduce the computational requirements without sacrificing accuracy or efficiency. 

An important advantage of the approach is that it allows the computation of the optimal significance level that can be used by the assurance monitor to ensure a bounded error 
rate while limiting the number of inputs for which an accurate prediction cannot be made. 
Unlike most common machine learning classifiers that assign a single label to an input, ICP computes a set of candidate labels that contains the correct class given a selected significance level. Small significance level values reduce the classification errors but may result in set predictors with multiple candidate labels. In autonomous systems, it is not only important to have predictions with well-calibrated confidence but also to be able to choose the desired significance level based on the application requirements.  Even though reducing the number of possible classes may be helpful when the information is provide to a human, in an autonomous system it is desirable that the prediction is unique. Therefore, we assume that set predictions that contain multiple classes lead to a rejection of the input and require human intervention. For this reason, it is desirable to minimize the number of test inputs with multiple predictions. If the prediction is unique, then the monitor ensures a confident prediction with well-calibrated error rate defined by the significance level. If the predicted set contains multiple predictions, the monitor rejects the prediction and raises an alarm. Finally, if the predicted set is empty the monitor indicates that no label is probable. We distinguish between multiple and no predictions, because they may lead to different action in the system. For example, no prediction may be the result of out-of-distribution inputs while multiple possible predictions may be an indication that the significance level is smaller than the accuracy of the underlying DNN. 

The paper presents a comprehensive empirical evaluation of the approach using three datasets for classification problems in CPS of increasing complexity. The first dataset is the SCITOS-G5 robot navigation dataset~\parencite{Dua:2019} for which we use a fully connected feedforward network architecture. The second is a speech recognition dataset which contains audio files of human speech~\parencite{speech_dataset}. For this problem, we learn the embedding representations using a DNN with 1D convolutional layers. The third dataset is the German Traffic Sign Recognition Benchmark (GTSRB)~\parencite{GTSRB_cite}. For this dataset, we use a modified version of the VGG16 architecture~\parencite{simonyan2014very} to learn and generate the embedding representations. We used different combinations of NC functions and distance metric learning architectures and compare them with ICP without distance metric learning. The results demonstrate that that the selected or computed significance levels bound the error-rate in all cases. Moreover, the representations learned by the siamese or triplet networks result in well-formed clusters for different classes and individual training data typically can be captured by their class centroid. Such representations reduce the memory requirements and the execution time overhead while still ensure a bounded small error-rate with limited number of prediction sets containing multiple candidate labels.

Related work on confidence estimation and well-calibrated models for different kind of machine learning methods is presented in Section~\ref{sec:related_work}. In Section~\ref{sec:problem}, we define the problem and present the proposed architecture. Sections~\ref{sec:distance_learning}-\ref{sec:assurance_monitoring} present the details of ICP based on distance learning and assurance monitoring.  Finally, we evaluate the performance of our suggested approach on three different applications in Section~\ref{sec:evaluation}.

\section{Related Work}
\label{sec:related_work}
Machine learning components tend to be poorly calibrated. Modern, commonly used DNN architectures typically have a softmax layer to produce a probability-like output for each class. The chosen class is the one with the highest probability, however this generated probability measure is often higher than the actual posterior probability that the prediction is correct. Other factors that affect the calibration in DNNs are the depth, width, weight decay and Batch Normalization~\parencite{Guo:2017:CMN:3305381.3305518}. The estimation of accurate error-rate bounds is important as it provide assurance guarantees in safety-critical applications  but also makes the decision confidence interpretable by humans. Several approaches have been proposed that compute well-calibrated confidence metrics in different ways, like scaling the DNN softmax outputs or other post-processing algorithms.

The calibration methods generally belong to two categories: parametric and  non-parametric. The parametric methods assume that the  probabilities follow certain well-known  distributions whose parameters are to be estimated from the training data. The Platt's scaling method~\parencite{Platt99probabilisticoutputs} is proposed for the calibration of Support Vector Machine (SVM) outputs. After the training of an SVM, the method computes the parameters of a sigmoid function to map the outputs into probabilities. Piecewise logistic regression is an extension of Platt scaling and assumes that the log-odds of calibrated probabilities follow a piecewise linear function~\parencite{zhang2004probabilistic}. Another variant of Platt scaling is temperature scaling~\parencite{Guo:2017:CMN:3305381.3305518} which can be applied in DNNs with  a softmax output layer. After training of a DNN, a temperature scaling factor $T$ is computed on a validation set to scale the softmax outputs. However, while temperature scaling achieves good calibration when the data in the validation dataset are independent and identically distributed (IID), there is no calibration guarantee under distribution shifts~\parencite{ovadia2019can}. Experiments in~\parencite{kumar2019verified} show that Platt scaling and temperature scaling are not as well calibrated as it is reported and it is difficult to know how miscalibrated they are.

Histogram binning or quantile binning is a commonly used non-parametric approach with either equal-width or equal-frequency bins. It divides the outputs of a classifier into bins and computes the calibrated probability as the ratio of correct classifications in each bin~\parencite{zadrozny2001obtaining}. Isotonic Regression is a generalization of histogram binning by jointly optimizing the bin boundaries and bin predictions~\parencite{Zadrozny:2002:TCS:775047.775151}. An extention of isotonic regression is a method called ensemble of near-isotonic regression (ENIR) that uses selective Bayesian averaging to ensemble the nearly-isotonic regression models~\parencite{naeini2018binary}. Adaptive calibration of predictions (ACP) also uses the ratio of correct classifications as the posterior probability in each bin, but it obtains bins from a 95\% confidence interval around each individual prediction~\parencite{jiang2012calibrating}. Estimating calibrated probability is a more significant issue in class imbalance and class overlap problems. Receiver  Operating  Characteristics (ROC) Binning uses the ROC curves to construct equal-width bins that provide accurate calibrated probabilities that are robust to changes in the prevalence of the positive class~\parencite{sun2018obtaining}. Bayesian binning into quantiles (BBQ) extends the simple histogram-binning  calibration method by considering multiple equal frequency Histogram Binning models and their combination as the calibration result~\parencite{naeini2015obtaining}.

Another framework developed to produce well-calibrated confidence values is the Conformal Prediction (CP)~\parencite{Vovk:2005:ALR:1062391,Shafer:2008:TCP:1390681.1390693,balasubramanian2014conformal}. The conformal prediction framework can be applied to produce calibrated confidence values with a variety of machine learning algorithms with slight modifications. Using CP together with machine learning models such as DNNs is computationally inefficient. In \parencite{papadopoulos2007conformal}, the authors suggest a modified version of the CP framework, Inductive Conformal Prediction (ICP) that has less computational overhead and they evaluate the results using DNNs as undelying model. Deep $k$-Nearest Neighbors (DkNN) is an approach based on ICP for classification problems that uses the activations from all the hidden layers of a neural network as features~\parencite{papernot2018deep}. The method is based on the assumption that when a DNN makes a wrong prediction, there is a specific hidden layer that generated intermediate results that lead to the wrong prediction. Taking into account all the hidden layers can lead to better interpretability of the predictions. In \parencite{johansson2013conformal}, the authors present an empirical investigation of decision trees as conformal predictors and analyzed the effects of different split criteria, such as the Gini index and the entropy, on ICP. There are similar evaluations using ICP with random forests \parencite{Bhattacharyya2013, 10.1007/978-3-642-16239-8_8} as well as SVMs \parencite{MAKILI20111213}. The above methods are applied to datasets and show good results when the input data are IID. In~\parencite{boursinos2020improving} we showed that ICP under-performs when the input data are sequential. Individual frames of a sequence might contain partial information regarding the input and more frames might be needed for ICP to reach a confident prediction. The performance of ICP in this case can be improved by designing a feedback-loop configuration that queries the sensors until a single confident decision can be reached.

Confidence bounds can also be generated for regression problems. In this case instead of sets of multiple candidate labels we have intervals around a point prediction that include the correct prediction with a desired confidence. There are ICP methods for regression problems with different underlying machine learning algorithms. In \parencite{papadopoulos2011regression}, the authors use the $k$-Nearest Neighbours Regression ($k$-NNR) as a predictor and evaluate the effects of different nonconformity functions. Random forests can also be used in regression problems. In \parencite{johansson2014regression}, there is a comparison on the generated confidence bounds using $k$-NNR and DNNs \parencite{papadopoulos2011reliable}. An alternative framework used to compute confidence bounds on regression problems is the Simultaneous Confidence Bands. The method presented in \parencite{10.2307/2242228} generates linear confidence bounds centered around the point prediction of a regression model. In this approach, the model used for predictions has to be estimated by a sum of linear models. Models that satisfy this condition are the least squares polynomial models, kernel methods and smoothing splines. Functional Principal Components (FPC) analysis can be used for the decomposition of an arbitrary regression model to a combination of linear models \parencite{goldsmith2013corrected}. 

The findings of the state-of-the-art methods described above illustrate the significance of computing well-calibrated and accurate confidence measures. 
Typically, the main objective is to complement existing machine learning models that are generally unable to produce an accurate estimation of confidence for their predictions with post-processing techniques in order to compute well-calibrated probabilities. 
An important advantage of such approaches is that they are independent of the underline predictive machine learning models. Therefore, there is no need to redesign and optimize the objective functions used for training which could lead to optimization tasks with high computational complexity.

Computing well-calibrated confidence is extremely important for designing autonomous systems because accurate measures of confidence are necessary to estimate the risk associated with 
each decision. The main limitation of existing methods comes from the fact that it is very difficult to select desired confidence values according to the application requirements and ensure bounded error-rate. This is especially important in autonomous CPS applications where decisions can be safety critical.
Another important challenge is to investigate how the computed confidence measures can be used for decision making by autonomous systems and how to handle data for which a confident decision cannot be taken.

The proposed work based on ICP produces prediction sets and computes a significance level that will bound the expected error-rate. Similar to existing methods, since the approach is based on ICP, it can be used with any machine learning component without the need of retraining. ICP methods provide very promising results especially when the input data are not very high-dimensional and there are not stringent time constraints. 
However, ICP can be impractical when the inputs are, for example, images because of the excessive memory requirements and high execution times. The proposed approach aims to learn appropriate lower-dimensional representations of high-dimensional inputs that make the task of computing confidence measures based on similarities much easier.

\section{Problem Formulation}
\label{sec:problem}

A perception component in a CPS aims to observe and interpret the environment in order to provide information for decision making.
For example, in autonomous vehicles a DNN can be used to classify traffic signs. 
The problem is to complement the prediction of the DNN
 with a computation of confidence. 
 An efficient and robust approach must ensure a small and well-calibrated error rate while limiting 
 the number of alarms to enable real-time monitoring.
 That is, maximize the autonomous operation time while keeping the error-rate bounded according to the application requirements. Finally, the computation of well-calibrated predictions must be computationally efficient for applications with high-dimensional inputs that require fast decision as, for example, in autonomous vehicles.

During the system operation of a CPS, inputs arrive one by one. After receiving each input,
the objective is to compute a valid measure of the confidence of the prediction.
The objective is twofold: (1) provide guarantees for the error rate of the prediction and 
(2) design a monitor which limits the number of input examples for which a confident prediction 
cannot be made. 
Such a monitor can be used, for example, by generating warnings that 
require human intervention.

\begin{figure}[ht]
\centering
% We need layers to draw the block diagram
\pgfdeclarelayer{background}
\pgfdeclarelayer{foreground}
\pgfsetlayers{background,main,foreground}

% Define a few styles and constants
\tikzstyle{environment}=[draw, fill=blue!30, text width=5em,  
text centered, minimum height=4em]
\tikzstyle{blocks}=[draw, fill=yellow!70, text width=5em, 
text centered, minimum height=4em]
\tikzstyle{data_in}=[draw, cylinder, shape border rotate=90, fill=orange!70, text width=5em, 
text centered, shape aspect=.4, minimum height=6.5em]

\begin{tikzpicture}
	\node (env) [environment, rounded corners=0.1cm, text width=6em] {Environment};
	\node [blocks, right of=env, node distance=5cm, align=center, rounded corners=0.1cm] (icp) {ICP};
	\node [blocks, right of=icp, node distance=4.5cm, align=center, rounded corners=0.1cm] (decision) {Assurance\\Monitoring};
	\node (system) [environment, rounded corners=0.1cm, text width=4em, right of=decision, node distance=2.5cm] {System};
	\node (distance) [data_in,above left=1cm and 0.1cm of icp] {Distance Learning DNN};
	\node (calib) [data_in,above right=1cm and 0.1cm of icp] {Calibration NC Scores};
	
	\draw[->,thick] (env)-- node[below,near end,text width=3em,align=center]{Sensor\\Inputs}(icp);
	\draw[->,thick] (icp)-- node[below,near start,text width=3.1em,align=center]{Prediction\\Sets}(decision);
	\draw[->,thick] (decision)-- (system);
	\draw[->,thick] (distance.south)--+(0,-0.5)-|($(icp.north west)!0.3!(icp.north east)$);
	\draw[->,thick] (calib.south)--+(0,-0.5)-|($(icp.north west)!0.7!(icp.north east)$);
	
	\begin{pgfonlayer}{background}
		% Compute a few helper coordinates
		\path (distance.west |- distance.north)+(-0.5,0.3) node (a) {};
		\path (icp.south -| calib.east)+(+0.3,-1) node (b) {};
		\path[fill=red!20!green!30,rounded corners=0.2cm, draw=black!50, dashed]
		(a) rectangle (b) node[pos=.5, yshift=-2.7cm, text width=10.0em] {};
	\end{pgfonlayer}
\end{tikzpicture}
\caption{Assurance monitoring using ICP based on distance learning.}
\label{fig:approach}
\end{figure}
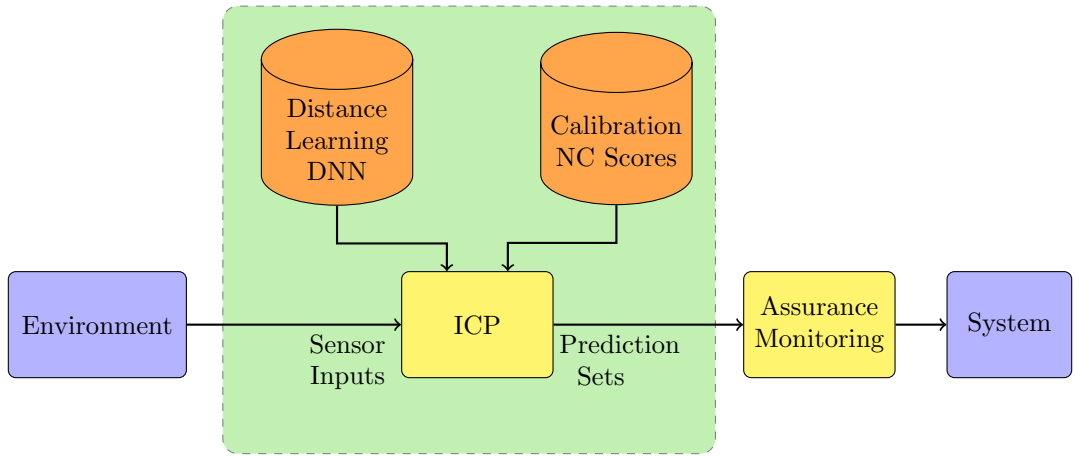

The conformal prediction framework allows computing set predictors for a given confidence expressed as a significance value~\parencite{balasubramanian2014conformal}. The confidence is generated by comparing how similar a test is to the training data using different nonconformity functions. In our previous work~\parencite{boursinos2020trusted} we used DNNs to produce embedding representations for more efficient application of ICP. The additional problem we are solving is the computation of appropriate embedding representations that will lead to more confident decisions. The proposed approach is illustrated in Figure~\ref{fig:approach}. The main idea is to use distance learning and enable DNNs to learn a lower-dimensional representation for each input on an embedding space where the Euclidean distance between the input representations is a measure of similarity between the original inputs themselves. The ICP approach is applied using the low-dimensional embedding representations and estimates the similarity between a new input and the available data in the training set using a NC function. 
Using such a representation not only reduces the execution time and the memory requirements but is also more efficient in producing useful predictions. Based on a chosen significance level, ICP generates a set of possible predictions. If the computed set contains a single prediction, the confidence is a well-calibrated and a valid indication of the expected 
error. If the computed set contains multiple predictions or no predictions, an alarm can be raised
to indicate the need for additional information.

In CPS, it is desirable to minimize the number of alarms while performing the required computations in real-time.
Evaluation of the method must be based on metrics that quantify the error rate, the number of alarms,
and the computational efficiency. For real-time operation, the time and memory requirements of the monitoring approach 
must be similar to the computational requirements of the DNNs used in the CPS architecture. 
Figure~\ref{fig:approach} illustrates the proposed architecture for assurance monitoring. At design time, a DNN is trained to produce embedding representations using distance metric learning techniques. Then,  NC scores are computed for a labeled calibration set that is not used for training of the DNN. During system operation, the assurance monitor employs the trained DNN to map new sensor inputs to lower-dimensional representations. Using the NC scores of the calibration data, the method produces prediction sets including well-calibrated confidence of the predictions. Ideally, a prediction set should include exactly one class to enable decision making. Alarms can be raised if either the prediction set include multiple possible classes or if it does not contain any.

\section{Distance Learning}
\label{sec:distance_learning}

\begin{figure}[ht]
\centering
\tikzstyle{NETS}=[draw, fill=blue!30, rounded corners=0.1cm, text centered, minimum height=1.2cm, minimum width=1.7cm]

\def\rsize{0.2}

\begin{tikzpicture}
	[axis/.style={thin, black, ->, >=stealth'}]
	
	\node[inner sep=0pt] (sign0) at (0,0)
	{\includegraphics[scale=0.3]{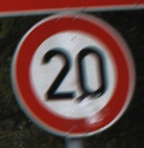}};
	\node[inner sep=0pt, below of=sign0, node distance=2cm] (sign1)
	{\includegraphics[scale=0.4]{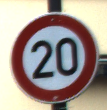}};
	\node[inner sep=0pt, below of=sign1, node distance=2cm] (sign2)
	{\includegraphics[scale=0.4]{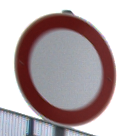}};
	
	\node[NETS, right of=sign0, node distance=2.5cm] (dnn0) {DNN};
	\node[NETS, right of=sign1, node distance=2.5cm] (dnn1) {DNN};
	\node[NETS, right of=sign2, node distance=2.5cm] (dnn2) {DNN};
	
	\draw[->,thick] ($(sign0.east)+(0.1,0)$)--($(dnn0.west)+(-0.1,0)$);
	\draw[->,thick] ($(sign1.east)+(0.1,0)$)--($(dnn1.west)+(-0.1,0)$);
	\draw[->,thick] ($(sign2.east)+(0.1,0)$)--($(dnn2.west)+(-0.1,0)$);
	
	\draw ($ (dnn0.east) + (1,0) $) -- ($ (dnn0.east) + (1+\rsize,0) $);
	\draw ($ (dnn1.east) + (1,0) $) -- ($ (dnn1.east) + (1+\rsize,0) $);
	\draw ($ (dnn2.east) + (1,0) $) -- ($ (dnn2.east) + (1+\rsize,0) $);
	
	\draw ($ (dnn0.east) + (1,0) $) -- ($ (dnn0.east) + (1,4*\rsize) $);
	\draw ($ (dnn0.east) + (1,0) $) -- ($ (dnn0.east) + (1,-4*\rsize) $);
	\draw ($ (dnn0.east) + (1+\rsize,0) $) -- ($ (dnn0.east) + (1+\rsize,4*\rsize) $);
	\draw ($ (dnn0.east) + (1+\rsize,0) $) -- ($ (dnn0.east) + (1+\rsize,-4*\rsize) $);
	
	\draw ($ (dnn1.east) + (1,0) $) -- ($ (dnn1.east) + (1,4*\rsize) $);
	\draw ($ (dnn1.east) + (1,0) $) -- ($ (dnn1.east) + (1,-4*\rsize) $);
	\draw ($ (dnn1.east) + (1+\rsize,0) $) -- ($ (dnn1.east) + (1+\rsize,4*\rsize) $);
	\draw ($ (dnn1.east) + (1+\rsize,0) $) -- ($ (dnn1.east) + (1+\rsize,-4*\rsize) $);
	
	\draw ($ (dnn2.east) + (1,0) $) -- ($ (dnn2.east) + (1,4*\rsize) $);
	\draw ($ (dnn2.east) + (1,0) $) -- ($ (dnn2.east) + (1,-4*\rsize) $);
	\draw ($ (dnn2.east) + (1+\rsize,0) $) -- ($ (dnn2.east) + (1+\rsize,4*\rsize) $);
	\draw ($ (dnn2.east) + (1+\rsize,0) $) -- ($ (dnn2.east) + (1+\rsize,-4*\rsize) $);
	
	\foreach \x in {1,...,4}
		\draw ($ (dnn0.east) + (1,-\x*\rsize) $) -- ($ (dnn0.east) + (1+\rsize,-\x*\rsize) $);
	\foreach \x in {1,...,4}
		\draw ($ (dnn0.east) + (1,\x*\rsize) $) -- ($ (dnn0.east) + (1+\rsize,\x*\rsize) $);
	\foreach \x in {1,...,4}
	\draw ($ (dnn1.east) + (1,-\x*\rsize) $) -- ($ (dnn1.east) + (1+\rsize,-\x*\rsize) $);
	\foreach \x in {1,...,4}
	\draw ($ (dnn1.east) + (1,\x*\rsize) $) -- ($ (dnn1.east) + (1+\rsize,\x*\rsize) $);
	\foreach \x in {1,...,4}
	\draw ($ (dnn2.east) + (1,-\x*\rsize) $) -- ($ (dnn2.east) + (1+\rsize,-\x*\rsize) $);
	\foreach \x in {1,...,4}
	\draw ($ (dnn2.east) + (1,\x*\rsize) $) -- ($ (dnn2.east) + (1+\rsize,\x*\rsize) $);
	
	\draw[->,thick] ($(dnn0.east)+(0.1,0)$)--($(dnn0.east)+(0.9,0)$);
	\draw[->,thick] ($(dnn1.east)+(0.1,0)$)--($(dnn1.east)+(0.9,0)$);
	\draw[->,thick] ($(dnn2.east)+(0.1,0)$)--($(dnn2.east)+(0.9,0)$);
	
	%AXIS
	\draw[axis] ($(dnn1.east)+(3,-0.5)$) -- ($(dnn1.east)+(3,2.9)$) node [above, black] {};
	\draw[axis] ($(dnn1.east)+(3,-0.5)$) -- ($(dnn1.east)+(6,-0.5)$) node [above, black] {};
	\draw[axis] ($(dnn1.east)+(3,-0.5)$) -- ($(dnn1.east)+(2,-2.5)$) node [above, black] {};
	
	\node[draw, fill=green!30!black!40,circle,inner sep=0pt,minimum size=5pt] (r0) at ($(dnn1.east)+(3.8,2)$) {};
	\node[draw, fill=green!30!black!40,circle,inner sep=0pt,minimum size=5pt] (r1) at ($(dnn1.east)+(4,1.8)$) {};
	\node[draw, fill=red!80!black!40,circle,inner sep=0pt,minimum size=5pt] (r2) at ($(dnn1.east)+(3.5,-0.9)$) {};
	
	\draw[->,dashed] ($(dnn0.east)+(1.1+\rsize,0)$) to [out=30,in=150] ($(r0.west)+(-0.1,0.1)$);
	\draw[->,dashed] ($(dnn1.east)+(1.1+\rsize,0)$) to [out=0,in=-120] ($(r1.west)+(0,-0.2)$);
	\draw[->,dashed] ($(dnn2.east)+(1.1+\rsize,0)$) to [out=50,in=170] ($(r2.west)+(-0.1,0)$);
	
	\node[below of=sign2,node distance=1.5cm] (orig) {Original Inputs};
	\node[right of=orig, node distance=6.5cm] (repr) {Embedding Representations};

\end{tikzpicture}
\caption{Embedding representations of input images from the traffic sign recognition dataset.}
\label{fig:embedding_reprsentations}
\end{figure}
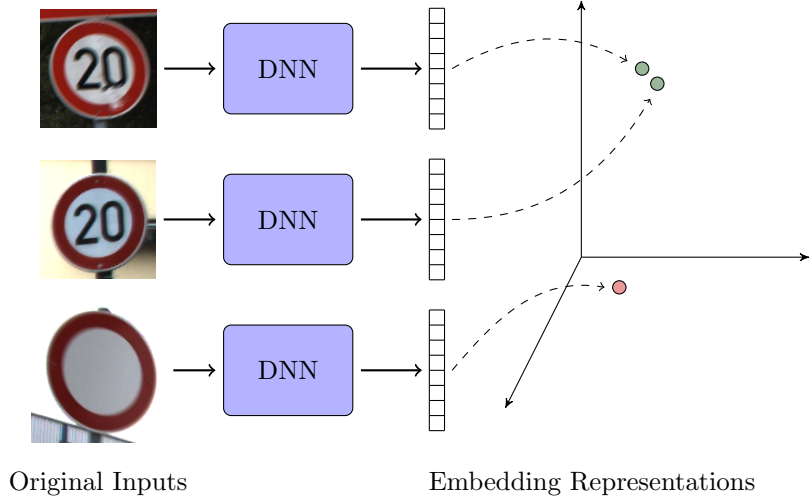

The ICP framework requires computing the similarity between the training data and a test input. This can be done efficiently by learning representations of the inputs for which the Euclidean distance is a metric of similarity, meaning that similar inputs will be close to each other as illustrated in Figure~\ref{fig:embedding_reprsentations}. There are different approaches based on DNN architectures that generate embedding representations for distance metric learning.

\begin{figure}[ht]
    \centering
    \begin{subfigure}{.5\textwidth}
        \centering
        \tikzstyle{NETS}=[draw, fill=blue!30, text width=4em, text centered, minimum height=2.5cm]
\tikzstyle{LOSS}=[draw, fill=yellow!70, text width=4em,  text centered, minimum height=1cm, minimum width=4cm]

\def\rsize{0.2}

\begin{tikzpicture}
	\node (net1) [NETS, rounded corners=0.1cm] {Net(x)};
	\node (net2) [NETS, right of=net1, node distance=3.4cm, rounded corners=0.1cm] {Net(x)};
	\node (a) at ($(net1)!0.5!(net2)$) {};
	\node (cont) [LOSS, rounded corners=0.1cm, above of=a, node distance=3.5cm, align=center] {Contrastive\\Loss};
	
	\draw[->,thick] (net1.south)+(0,-1cm) -- node[below, yshift=-0.5cm] {$x_1$} (net1.south);
	\draw[->,thick] (net2.south)+(0,-1cm) -- node[below, yshift=-0.5cm] {$x_2$} (net2.south);
	\draw[<->,thick] (net1.east) -- node[below, midway] {\scriptsize Weights}  (net2.west);
	
	\draw[->,thick] (net1.north) -- +(0,0.5cm);
	\draw[->,thick] (net2.north) -- +(0,0.5);
	
	\draw ($ (net1.north) + (0,0.5) $) -- ($ (net1.north) + (4*\rsize,0.5) $);
	\draw ($ (net1.north) + (0,0.5) $) -- ($ (net1.north) + (-4*\rsize,0.5) $);
	\draw ($ (net1.north) + (0,0.5+\rsize) $) -- ($ (net1.north) + (4*\rsize,0.5+\rsize) $);
	\draw ($ (net1.north) + (0,0.5+\rsize) $) -- ($ (net1.north) + (-4*\rsize,0.5+\rsize) $);
	\draw ($ (net2.north) + (0,0.5) $) -- ($ (net2.north) + (4*\rsize,0.5) $);
	\draw ($ (net2.north) + (0,0.5) $) -- ($ (net2.north) + (-4*\rsize,0.5) $);
	\draw ($ (net2.north) + (0,0.5+\rsize) $) -- ($ (net2.north) + (4*\rsize,0.5+\rsize) $);
	\draw ($ (net2.north) + (0,0.5+\rsize) $) -- ($ (net2.north) + (-4*\rsize,0.5+\rsize) $);
	
	\draw ($ (net1.north) + (0,0.5) $) -- ($ (net1.north) + (0,0.5+\rsize) $);
	\draw ($ (net2.north) + (0,0.5) $) -- ($ (net2.north) + (0,0.5+\rsize) $);
	\foreach \x in {1,...,4}
	\draw ($ (net1.north) + (-\x*\rsize,0.5) $) -- ($ (net1.north) + (-\x*\rsize,0.5+\rsize) $);
	\foreach \x in {1,...,4}
	\draw ($ (net1.north) + (\x*\rsize,0.5) $) -- ($ (net1.north) + (\x*\rsize,0.5+\rsize) $);
	\foreach \x in {1,...,4}
	\draw ($ (net2.north) + (-\x*\rsize,0.5) $) -- ($ (net2.north) + (-\x*\rsize,0.5+\rsize) $);
	\foreach \x in {1,...,4}
	\draw ($ (net2.north) + (\x*\rsize,0.5) $) -- ($ (net2.north) + (\x*\rsize,0.5+\rsize) $);
	
	\node (r1) at ($(net1.north)+(1.1,0.6)$) {$r_1$};
	\node (r2) at ($(net2.north)+(1.1,0.6)$) {$r_2$};
	
	\draw[->,thick] ($ (net1.north) + (0,0.5+\rsize) $) -- +(0,0.5) -| ($(cont.south east)!0.7!(cont.south west)$);
	\draw[->,thick] ($ (net2.north) + (0,0.5+\rsize) $) -- +(0,0.5) -| ($(cont.south east)!0.3!(cont.south west)$);
\end{tikzpicture}
        \caption{}
        \label{fig:siamese_network}
    \end{subfigure}%
    \begin{subfigure}{.5\textwidth}
        \centering
        \tikzstyle{NETS}=[draw, fill=blue!30, text width=4em, text centered, minimum height=2.5cm]
\tikzstyle{LOSS}=[draw, fill=yellow!70, text centered, minimum height=1cm, minimum width=5cm]

\def\rsize{0.2}

\begin{tikzpicture}
	\node (net1) [NETS, rounded corners=0.1cm] {Net(x)};
	\node (net2) [NETS, right of=net1, node distance=2.8cm, rounded corners=0.1cm] {Net(x)};
	\node (net3) [NETS, right of=net2, node distance=2.8cm, rounded corners=0.1cm] {Net(x)};
	\node (cont) [LOSS, rounded corners=0.1cm, above of=net2, node distance=3.5cm, align=center] {Triplet Loss};
	
	\draw[->,thick] (net1.south)+(0,-1cm) -- node[below, yshift=-0.4cm] {$x^-$} (net1.south);
	\draw[->,thick] (net2.south)+(0,-1cm) -- node[below, yshift=-0.5cm] {$x$} (net2.south);
	\draw[->,thick] (net3.south)+(0,-1cm) -- node[below, yshift=-0.4cm] {$x^+$} (net3.south);
	\draw[<->,thick] (net1.east) -- node[below, midway] {\scriptsize Weights}  (net2.west);
	\draw[<->,thick] (net2.east) -- node[below, midway] {\scriptsize Weights}  (net3.west);
	
	\draw[->,thick] (net1.north) -- +(0,0.5cm);
	\draw[->,thick] (net2.north) -- +(0,0.5cm);
	\draw[->,thick] (net3.north) -- +(0,0.5cm);
	
	\draw ($ (net1.north) + (0,0.5) $) -- ($ (net1.north) + (4*\rsize,0.5) $);
	\draw ($ (net1.north) + (0,0.5) $) -- ($ (net1.north) + (-4*\rsize,0.5) $);
	\draw ($ (net1.north) + (0,0.5+\rsize) $) -- ($ (net1.north) + (4*\rsize,0.5+\rsize) $);
	\draw ($ (net1.north) + (0,0.5+\rsize) $) -- ($ (net1.north) + (-4*\rsize,0.5+\rsize) $);
	\draw ($ (net2.north) + (0,0.5) $) -- ($ (net2.north) + (4*\rsize,0.5) $);
	\draw ($ (net2.north) + (0,0.5) $) -- ($ (net2.north) + (-4*\rsize,0.5) $);
	\draw ($ (net2.north) + (0,0.5+\rsize) $) -- ($ (net2.north) + (4*\rsize,0.5+\rsize) $);
	\draw ($ (net2.north) + (0,0.5+\rsize) $) -- ($ (net2.north) + (-4*\rsize,0.5+\rsize) $);
	
	\draw ($ (net3.north) + (0,0.5) $) -- ($ (net3.north) + (4*\rsize,0.5) $);
	\draw ($ (net3.north) + (0,0.5) $) -- ($ (net3.north) + (-4*\rsize,0.5) $);
	\draw ($ (net3.north) + (0,0.5+\rsize) $) -- ($ (net3.north) + (4*\rsize,0.5+\rsize) $);
	\draw ($ (net3.north) + (0,0.5+\rsize) $) -- ($ (net3.north) + (-4*\rsize,0.5+\rsize) $);
	
	\draw ($ (net1.north) + (0,0.5) $) -- ($ (net1.north) + (0,0.5+\rsize) $);
	\draw ($ (net2.north) + (0,0.5) $) -- ($ (net2.north) + (0,0.5+\rsize) $);
	\draw ($ (net3.north) + (0,0.5) $) -- ($ (net3.north) + (0,0.5+\rsize) $);
	\foreach \x in {1,...,4}
	\draw ($ (net1.north) + (-\x*\rsize,0.5) $) -- ($ (net1.north) + (-\x*\rsize,0.5+\rsize) $);
	\foreach \x in {1,...,4}
	\draw ($ (net1.north) + (\x*\rsize,0.5) $) -- ($ (net1.north) + (\x*\rsize,0.5+\rsize) $);
	\foreach \x in {1,...,4}
	\draw ($ (net2.north) + (-\x*\rsize,0.5) $) -- ($ (net2.north) + (-\x*\rsize,0.5+\rsize) $);
	\foreach \x in {1,...,4}
	\draw ($ (net2.north) + (\x*\rsize,0.5) $) -- ($ (net2.north) + (\x*\rsize,0.5+\rsize) $);
	\foreach \x in {1,...,4}
	\draw ($ (net3.north) + (-\x*\rsize,0.5) $) -- ($ (net3.north) + (-\x*\rsize,0.5+\rsize) $);
	\foreach \x in {1,...,4}
	\draw ($ (net3.north) + (\x*\rsize,0.5) $) -- ($ (net3.north) + (\x*\rsize,0.5+\rsize) $);
	
	\node (r1) at ($(net1.north)+(1.1,0.6)$) {$r^-$};
	\node (r2) at ($(net2.north)+(1.1,0.6)$) {$r$};
	\node (r3) at ($(net3.north)+(1.1,0.6)$) {$r^+$};
	
	\draw[->,thick] ($ (net1.north) + (0,0.5+\rsize) $) -- +(0,0.5) -| ($(cont.south east)!0.8!(cont.south west)$);
	\draw[->,thick] (net2)--(cont);
	\draw[->,thick] ($ (net3.north) + (0,0.5+\rsize) $) -- +(0,0.5) -| ($(cont.south east)!0.2!(cont.south west)$);
\end{tikzpicture}     
        \caption{}
        \label{fig:triplet_network}
    \end{subfigure}
    
    \caption{(a) Siamese network architecture and (b) Triplet network architecture}
    \label{fig:distance_learning_architectures}
\end{figure}
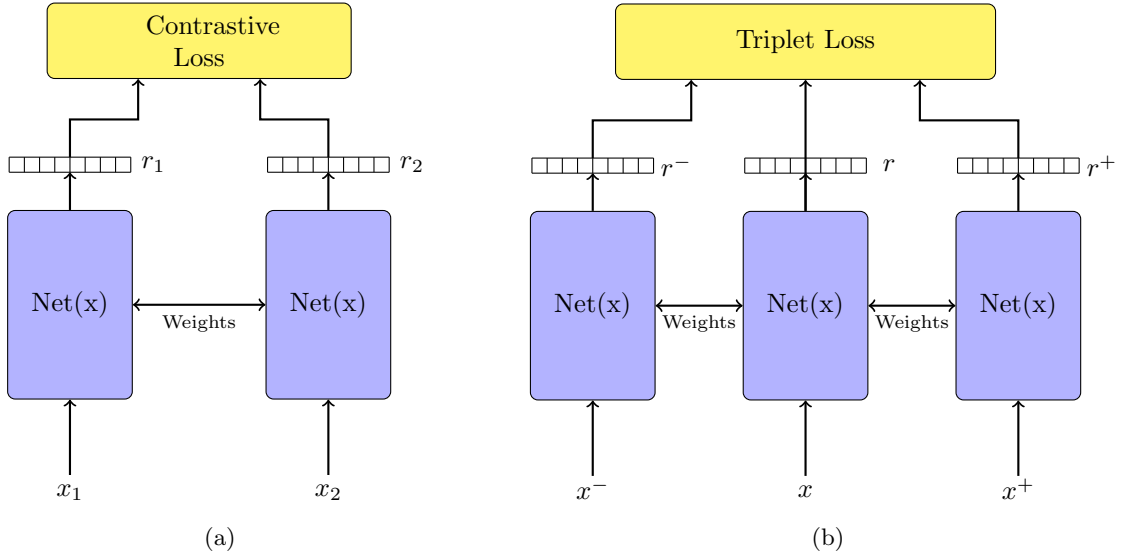

%siamese
A \textit{siamese network} is composed using two copies of the same neural network with shared parameters~\parencite{koch2015siamese} as shown in Figure~\ref{fig:siamese_network}. During training, each identical copy of the siamese network is fed with different training samples $x_1$ and $x_2$ belonging to classes $y_1$ and $y_2$. The embedding representations produced by each network copy are $r_1=\text{Net}(x_1)$ and $r_2=\text{Net}(x_2)$. The learning goal is to minimize the Euclidean distance between the embedding representations of inputs belonging to the same class and maximize it for inputs belonging to different classes as described below:
\begin{equation}
\label{eq:optimization_problem}
% d(x_1,x_2)= 
\begin{cases}
    \min d(r_1,r_2),& \text{if } y_1=y_2\\
    \max d(r_1,r_2),& \text{otherwise}
\end{cases}
\end{equation}
This optimization problem can be solved using the \textit{contrastive loss} function~\parencite{melekhov2016siamese}:
$$L(r_1,r_2,y)=y \cdot d(r1,r2)+(1-y)\max[0,m-d(r_1,r_2)]$$
where $y$ is a binary flag equal to 0 if $y_1=y_2$ and to 1 if $y_1\neq y_2$ and $m$ is a margin parameter. In particular, when $y_1\neq y_2$, $L=0$ when $d(r1,r2)\geq m$, otherwise the parameters of the network are updated to produce more distant representations for those two elements. The reason behind the use of the margin is that when the distance between pairs of different classes are large enough and at most $m$, there is no reason to update the network to put the representations even further away from each other and instead focus the training on harder examples.

%triplet
Another architecture trained to produce embedding representations for distance learning is the \textit{triplet network}~\parencite{hoffer2014deep}. A triplet is composed using three copies of the same neural network with shared parameters as shown in Figure~\ref{fig:triplet_network}. The training examples consist of three samples, the anchor sample $x$, the positive sample $x^+$ and the negative sample $x^-$. The samples $x$ and $x^+$ belong to the same class while $x^-$ belongs to a different class. The embedding representations produced by each network copy will be $r=\text{Net}(x)$, $r^+=\text{Net}(x^+)$ and $r^-=\text{Net}(x^-)$. The optimization problem described by the equations \ref{eq:optimization_problem} is solved by training the triplet network copies using the \textit{triplet loss} function:

$$L(r,r^+,r^-)=\max[d(r,r^+)-d(r,r^-)+m,0]$$.

The margin parameter $m$ separates pairs of different classes by at most $m$ and it is used so that the network parameters will not be updated trying to push a pair even further away when a positive sample is already at least $m$ closer to an anchor than a negative sample.
Instead, the training is more efficient when harder triplets are used.
The input triplets to the network copies can be sampled randomly from the training data. However, as training progresses it is harder to randomly find triplets that produce $L(r,r^+,r^-)>0$ that will update the triplet network parameters. This leads to slow training and underfitted models. The training can be improved by carefully mining the training data that produce a large loss~\parencite{xuan2019improved}. For each training iteration, first, the anchor training data are randomly chosen. For each anchor, the hardest positive sample is chosen, meaning a sample from the same class as the anchor that is located the furthest away from the anchor. Then, the triplets are formed by mining hard negative samples that satisfy $d(r,r^-)<d(r,r^+)$ or semi-hard negatives that satisfy $d(r,r^-)<d(r,r^+)+m$. This way the formed triplet batches will produce gradients to update the shared weights between the DNN copies.

\section{ICP Based on Distance Learning}
\label{sec:ICP}
We consider a training set \{$z_1,\dots,z_l$\} of examples, where each $z_i\in Z$ is a pair $(x_i,y_i)$ with $x_i$ the feature vector and $y_i$ the label of that example. For a given unlabeled input $x_{l+1}$ and a chosen significance level $\epsilon$, the task is to compute a prediction set $\Gamma^\epsilon$ for which $P(y_{l+1}\notin\Gamma^\epsilon)<\epsilon$, where $y_{l+1}$ the ground truth label of the input $x_{l+1}$. ICP computes well-calibrated prediction sets with the underlying assumption that all examples ($x_i,y_i$), $i=1,2,\dots$ are independent and identically distributed (IID) generated from the same but typically unknown probability distribution.

Central to the application of ICP is a \textit{nonconformity function} or nonconformity measure (NCM) which shows how different a labeled input is from the examples in the training set. For a given test example $z_i$ with candidate label $\tilde{y}_i$, a NC function assigns a numerical score indicating how different the example $z_i$ is from the examples in  $\{z_1,\dots,z_{i-1},z_{i+1},\dots,z_n\}$. There are many possible NC functions that can be used~\parencite{balasubramanian2014conformal,Vovk:2005:ALR:1062391,Shafer:2008:TCP:1390681.1390693,model_agnostic,boursinos2020assurance}. For example, a NC function can be defined as the number of the $k$-nearest neighbors to $z_{l+1}$ in the training set with label different than the candidate label $\tilde{y}_{l+1}$  ($k$-nearest neighbors NCM). The input space is often high-dimensional which makes storing the whole training set impractical and the computation of the NC scores inefficient. To address this challenge, the proposed approach leverages distance metric learning methods to learn representations that enable applying ICP in real-time.

Nonconformity functions that can be defined in the embedding space learned by siamese and triplet networks are (1) the \textit{k-Nearest Neighbors} ($k$-NN)~\parencite{papernot2018deep}, (2) the \textit{one Nearest Neighbor} (1-NN)~\parencite{Vovk:2005:ALR:1062391}, and (3) the \textit{Nearest Centroid}~\parencite{balasubramanian2014conformal}. The $k$-NN NCM finds the $k$ most similar examples of a test input $x$ in the training data and counts how many of those are labeled different than the candidate label $y$. We denote $f:~X\rightarrow~V$ the mapping from the input space $X$ to the embedding space $V$ defined by either a siamese or a triplet network. Using the trained neural network, the encodings $v_i=f(x_i)$ are computed and stored for all the training data $ x_i $. Given a test input $x$ with encoding $v=f(x)$, we compute the $k$-nearest neighbors in $V$ and store their labels in a multi-set $\Omega$. The $k$-NN NCM of input $ x $ with a candidate label $y$ is defined as
\begin{equation*}
\label{eq:nonconformity_multiple_neighbors}
\alpha(x,y)=|i\in\Omega:i\neq y|.
\end{equation*}

The 1-NN NCM requires to find the most similar example of a test input $x$ in the training set that is labeled the same as the candidate label $y$ as well as the most similar example in the training set that belongs to any class other than $y$ and is defined as
\begin{equation*}
\label{eq:nonconformity_1neighbor}
\alpha(x,y)=\dfrac{\min_{i=1,\dots,n:y_i=y}d(v,v_i)}{\min_{i=1,\dots,n:y_i\neq y}d(v,v_i)}
\end{equation*}
where $ v = f(x) $, $v_i=f(x_i)$, and $d$ is the Euclidean distance metric in the $V$ space.

The Nearest Centroid NCM simplifies the task of computing individual training examples that are similar to a test input when there is a large amount of training data. We expect examples that belong to a particular class to be close to each other in the embedding space so for each class $y_i$ we compute its centroid $\mu_{y_i}=\frac{\sum_{j=1}^{n_i}v_j^i}{n_i}$, where $v_j^i$ is the embedding representation of the  $ j^{th} $ training example from class $y_i$ and $n_i$ is the number of training examples in class $y_i$. The NC function is then defined as
\begin{equation*}
\label{eq:nonconformity_nearest_centroid}
\alpha(x,y)=\dfrac{d(\mu_y,v)}{\min_{i=1,\dots,n:y_i\neq y}d(\mu_{y_i},v)}
\end{equation*}
where $ v = f(x) $. It should be noted that for computing the nearest centroid NCM, we need to store only the centroid for each class.

The NC score is an indication of how uncommon a test input is compared to the training data. Input data that come from the same distribution as the training data will produce low NC scores and are expected to lead to more confident classifications while unusual inputs will have higher NC score. However, this measure does not provide clear confidence information by itself but it can be used by comparing it with NCM scores computed using a \textit{calibration set} of known labeled data. Consider the training set \{$z_1,\dots,z_l$\}. This set is split into two parts, the proper training set \{$z_1,\dots,z_m$\} of size $m<l$ that will also be used for the training of the siamese or triplet network and the calibration set \{$z_{m+1},\dots,z_l$\} of size $l-m$. The NC scores $a(x_i,y_i)$, $i=m+1,\dots,l$, of the examples in the calibration set are computed and stored before applying the online monitoring algorithm. Given a test input $x$ with an unknown label $y$, the method generates a set $|\Gamma^\epsilon|$ of possible labels $\tilde{y}$ so that $P(y\notin|\Gamma^\epsilon|)<\epsilon$. For all the candidate labels $\tilde{y}$, ICP computes the empirical $p$-value defined as
\begin{equation*}
\label{eq:p_values_equation}
p_j(x)=\frac{|\{\alpha\in A:\alpha\geq\alpha(x,j)\}|}{|A|}.
\end{equation*}
which is the fraction of NC scores of the calibration data that are equal or larger than the NC score of a test input. A candidate label is added to $\Gamma^\epsilon$ if $ p_j(x) > \epsilon $. It is shown in~\parencite{balasubramanian2014conformal} that the prediction sets computed by ICP are valid, that is the probability of error will not exceed $ \epsilon $ for any $\epsilon \in [0,1] $ for any choice of NC function. Our approach focuses on computing small prediction sets in an efficient manner that allow assurance monitoring approach in real-time.

\section{Assurance Monitoring}
\label{sec:assurance_monitoring}
In CPS, it is not only important to have predictions with well-calibrated confidence but also to be able to choose the desired significance level based on the application requirements. ICP computes a prediction set $\Gamma^\epsilon$ with a chosen significance level $ \epsilon $ and $\Gamma^\epsilon$ may include any subset of all possible classes. Even though reducing the number of possible classes may be helpful when the information is provided to a human, in an autonomous system it is desirable that the prediction is unique, i.e., $|\Gamma^\epsilon|=1$. Therefore, we assume that set predictions that contain multiple classes, i.e., $|\Gamma^\epsilon| > 1$, lead to a rejection of the input and require human intervention. For this reason, it is desirable to minimize the number of test inputs with multiple predictions and we define a monitor with output defined as
\begin{equation*}
    out =
        \begin{cases}
            0,& \text{if } |\Gamma^\epsilon|=0\\
            1,& \text{if } |\Gamma^\epsilon|=1\\
            \text{reject},& \text{if } |\Gamma^\epsilon|>1
        \end{cases}.
\end{equation*}

If the set $\Gamma^\epsilon$ contains a single prediction, the monitor outputs $ out = 1 $ to indicate a confident prediction with well-calibrated error rate $ \epsilon $. If the predicted set contains multiple predictions, the monitor rejects the prediction and raises an alarm. Finally, if the predicted set is empty the monitor outputs $ out = 0 $ to indicate that no label is probable. We distinguish between multiple and no predictions, because they may lead to a different action in the system. For example, no prediction may be the result of out-of-distribution inputs while multiple possible predictions may be an indication that the significance level is smaller than the accuracy of the underlying DNN. Choosing a relatively small significance level that can consistently produce prediction sets with only one class is important. To do this, we apply ICP on the data in the calibration/validation set and compute the smallest significance level $\epsilon$ that does not produce any prediction set with $|\Gamma^\epsilon|>1$. Assuming the distribution of the test set is the same as the one of the calibration/validation set we expect the same value of $\epsilon$ to minimize the prediction sets with multiple classes on the test data.

The assurance monitoring approach is illustrated by Algorithm~\ref{alg:offline}~and~\ref{alg:online}. Algorithm~\ref{alg:offline} shows the tasks that need to performed at design time where first a distance metric learning network $f$ is trained using the proper training set $(X,Y)$ so that the computed embedding representations will form clusters for each class. Then, using the calibration data, both the NC scores $A$ and the optimal significance level $\epsilon$ are computed and stored. Algorithm~\ref{alg:online} shows the tasks that are performed at runtime for a sensor input $x_t$. The input first needs to be mapped to its embedding representation $v_t$. Then using the same NC function that is used for the calibration data, we compute the NC scores and the p-values assuming every label $j$ as candidate label. Then the p-values $p_j$ and $\epsilon$ are used to compute the set of candidate labels $\Gamma^\epsilon$.

\begin{algorithm}[H]
	\caption{\textbf{-- Training, Calibration and Significance Level computation}.}
	\label{alg:offline}
	\begin{algorithmic}[1] 
		\Require training data $(X, Y)$, calibration data $(X^c, Y^c)$
		\Require DNN architecture $f$ for distance metric learning
		\Require Nonconformity function $\alpha$
		\State Train $f$ using $(x, y) \in (X, Y)$
		\Comment{Training}
		\State // Compute the representations
		\State $V=f(X)$
		\State $V^c=f(X^c)$
		\State // Compute the nonconformity scores for the calibration data
		\State $A=\{\alpha(v^c,y^c) : (v^c, y^c) \in (V^c, Y^c)\}$ \Comment{Calibration}
		\For{each $v^c_i \text{ in } V^c, i=1..l-m$}
		\For{each label $j\in 1..n$}
		\State Compute the nonconformity score $\alpha(v^c_i,j)$
		\State $p_{ij}= p_j(v^c_i) = \frac{\left| \{ \alpha \in A : \alpha \geq \alpha(v^c_i, j) \} \right|}{|A|}$ \Comment{empirical $p$-value}
		\EndFor
		\State Store all $p_{ij}$
		\EndFor
		\State Compute $\epsilon$ such that for each $i \in [1..l-m]$ no more that than 1 of the p-values $p_{ij}\geq\epsilon$
% 		\Comment{$\epsilon$ computation}
	\end{algorithmic}
\end{algorithm}

\begin{algorithm}[H]
	\caption{\textbf{-- Assurance Monitoring}.}
	\label{alg:online}
	\begin{algorithmic}[1] 
	    \Require Nonconformity function $\alpha$
		\Require training data or centroids $(V, Y)$ depending on the used nonconformity function $\alpha$
		\Require trained siamese or triplet neural network $f$ for distance metric learning
		\Require test input $z_t=(x_t,y_t)$
		\Require significance level threshold $\epsilon$
		\State // Generate prediction sets for each test data $x_t$ 
		\State Compute embedding representation $v_t=f(x_t)$
		\For{each label $j\in 1..n$}
		\State Compute the nonconformity score $\alpha(v_t,j)$
		\State $p_j(z_t) = \frac{\left| \{ \alpha \in A : \alpha \geq \alpha(z, j) \} \right|}{|A|}$ \Comment{empirical $p$-value}
		\If{$p_j(z)\geq\epsilon$}
		\State Add $j$ to the prediction set $\Gamma^\epsilon$
		\Comment{$\Gamma^\epsilon$ formation}
		\EndIf
		\EndFor
		\If{$|\Gamma^\epsilon|=0$}
		\State \Return 0
		\ElsIf{$|\Gamma^\epsilon|=1$}
		\State \Return 1
		\Else
		\State\Return Reject
		\EndIf
	\end{algorithmic}
\end{algorithm}

\section{Evaluation}
\label{sec:evaluation}
Our assurance monitor design leverages distance metric learning techniques to compress the input data to lower dimensions in order to make the ICP application more efficient and with lower memory requirements. The objective of the evaluation is to compare how the suggested architecture performs against the baseline ICP approaches as well as investigate the validity/calibration and efficiency~(size of set predictions).

\subsection{Experimental Setup}
For the evaluation, we experiment with 3 datasets of variable complexity and input size. First, we use a dataset generated by the SCITOS-G5 mobile robot~\parencite{Dua:2019}. This robot is equipped with 24 ultrasound sensors around it that are sampled at a rate of 9 samples per second. Its task is to navigate itself around a room counter-clockwise in close proximity to the walls. The possible actions the robot can take to accomplish this are "Move-Forward", "Sharp-Right-Turn", "Slight-Left-Turn", and "Slight-Right-Turn". The SCITOS-G5 dataset contains 5456 raw values of the ultrasound sensor measurements as well as the decision it took in each sample. Because of the small sensor number, the inputs have one dimension and their size is relatively small. Second, we use a speech recognition dataset which contains 7501 audio samples from speeches of five prominent leaders; Benjamin Netanyahu, Jens Stoltenberg, Julia Gillard, Margaret Thatcher and Nelson Mandela, made available by the American Rhetoric~\parencite{speech_dataset}. Each audio sample has 1 second duration, the sampling rate is 16kHz and use Pulse-code modulation (PCM) encoding. Third, the German Traffic Sign Recognition Benchmark (GTSRB)
dataset is a collection of traffic sign images to be classified in 43 classes (each class corresponds to a type of traffic sign)~\parencite{GTSRB_cite}. The dataset has 26640 labeled images of various sizes between 15x15 to 250x250 depending on the distance of the traffic sign to the vehicle. For all datasets we split the available data so that 10\% of the samples is used for testing. From the remaining 90\% of the data, 80\% is used for training and 20\% for calibration and/or validation. In the ICP implementations that use the $k$-NN NC function the number of neighbors $k$ are chosen to be 20, 15 and 40 respectively for the 3 datasets, values that produce stability to outlier data points. The choice of DNN architectures happened according to the complexity of each application so that they will be simple enough to reduce the computational requirements but at the same time achieve good accuracy and data clustering without overfitting. All the experiments run in a desktop computer equipped with Intel(R) Core(TM) i9-9900K CPU, 32 GB RAM and a Geforce RTX 2080 GPU with 8 GB memory.

\subsection{Baseline}
The proposed approach assigns the original inputs into embedding representations for which the Euclidean distance is a measure of similarity between the inputs themselves. In order to understand the effect of the distance metric learning in ICP we compare it with the approaches we used in our previous work~\parencite{boursinos2020assurance}. First, the most basic way of applying ICP is using only the original inputs. Then we compare it with the approach we presented in our previous work that uses embedding representations without distance metric learning and this will be the baseline in the following experiments.

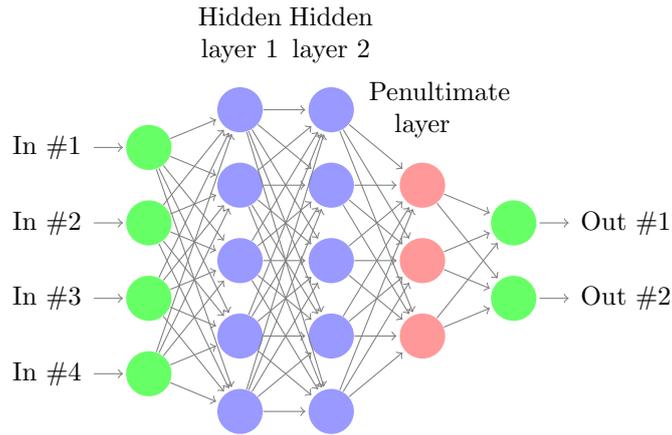
\begin{figure}[H]
\centering
\def\layersep{1.2cm} %separation between layers
\def\ha{4} %input neurons
\def\hb{5} %hidden 1
\def\hc{5} %hidden 2
\def\hd{3} %embedding
\def\ho{2} %output
\begin{tikzpicture}[shorten >=1pt,->,draw=black!50, node distance=\layersep]
    \tikzstyle{every pin edge}=[<-,shorten <=1pt]
    \tikzstyle{neuron}=[circle,fill=black!25,minimum size=17pt,inner sep=0pt]
    \tikzstyle{input neuron}=[neuron, fill=green!60];
    \tikzstyle{output neuron}=[neuron, fill=green!60];
    \tikzstyle{hidden neuron}=[neuron, fill=blue!40];
    \tikzstyle{embedding neuron}=[neuron, fill=red!40];
    \tikzstyle{annot} = [text width=4em, text centered]

    % Draw the input layer nodes
    \foreach \name / \y in {1,...,\ha}
    % This is the same as writing \foreach \name / \y in {1/1,2/2,3/3,4/4}
        \node[input neuron, pin=left:In \#\y] (I-\name) at (0,-\y) {};

    % Draw the hidden layer 1 nodes
    \foreach \name / \y in {1,...,\hb}
        \path[yshift=0.5cm]
            node[hidden neuron] (H1-\name) at (\layersep,-\y cm) {};
            
    % Draw the hidden layer 2 nodes
    \foreach \name / \y in {1,...,\hc}
        \path[yshift=0.5cm]
            node[hidden neuron] (H2-\name) at (2*\layersep,-\y cm) {};
            
    % Draw the embedding nodes
    \foreach \name / \y in {1,...,\hd}
        \path[yshift=-0.5cm]
            node[embedding neuron] (E-\name) at (3*\layersep,-\y cm) {};
            
    \foreach \name / \y in {1,...,\ho}
        \path[yshift=-1.0cm]
            node[output neuron, pin={[pin edge={->}]right:Out \#\y}] (O-\name) at (4*\layersep,-\y) {};

    % Draw the output layer node
    % \node[output neuron,pin={[pin edge={->}]right:Output}, right of=E-3] (O) {};

    % Connect every node in the input layer with every node in the
    % hidden layer.
    \foreach \source in {1,...,\ha}
        \foreach \dest in {1,...,\hb}
            \path (I-\source) edge (H1-\dest);
            
    % Connect every node in the hidden layer 1 with every node in the
    % hidden layer 2.
    \foreach \source in {1,...,\hb}
        \foreach \dest in {1,...,\hc}
            \path (H1-\source) edge (H2-\dest);
            
    % Connect every node in the hidden layer 2 with every node in the
    % embedding layer.
    \foreach \source in {1,...,\hc}
        \foreach \dest in {1,...,\hd}
            \path (H2-\source) edge (E-\dest);
            
    \foreach \source in {1,...,\hd}
        \foreach \dest in {1,...,\ho}
            \path (E-\source) edge (O-\dest);

    % Connect every node in the hidden layer 2 with the output layer
    % \foreach \source in {1,...,\hd}
    %     \path (E-\source) edge (O);

    % Annotate the layers
    \node[annot,above of=H1-1, node distance=1cm] (hl) {Hidden layer 1};
    \node[annot,above of=H2-1, node distance=1cm] (hl) {Hidden layer 2};
    \node[annot,above of=E-1, node distance=1cm] (hl) {Penultimate layer};

\end{tikzpicture}
\caption{Baseline DNN architecture}
\label{fig:baseline_architecture}
\end{figure}

The baseline approach computes the embedding representations using the activations of the penultimate layer of a DNN. A DNN is trained as a classifier to predict the class of the input data. The vector of activations of the neurons in the penultimate layer will be considered as the embedding representation of the input. In Figure~\ref{fig:baseline_architecture} there is an illustration of how the embedding representations are generated in the baseline using a DNN with 4 input neurons that classify inputs to two possible classes. The embedding representations are generated in the penultimate layer and are typically reduced in size compared to the inputs. For an accurate comparison between the baseline and the proposed improvements using either the triplet or the siamese network all of these approaches use the same DNN architecture, meaning the embedding representations will also be of the same size.

\subsection{Preprocessing and Distance Learning}
The difficulty to compute the NC functions and the memory demands increase as the input size increases. Here we see how the original high-dimensional inputs are mapped to lower-dimensional representations
so that the application of ICP will be more efficient as well as the Euclidean distance between two embedding representations is a metric of similarity, property useful in the computation of the NC scores. We evaluate how the use of the embedding representations affect the application of ICP when it is applied on datasets of increasing complexity. First, the input data to the SCITOS-G5 mobile robot is a vector of 24 values. We use a fully connected feedforward DNN to generate embedding representations with size 8. The DNN is trained in either a siamese or triplet network for distance metric learning. The triplet network is trained without mining since this is a small dataset. Second, the speech recognition dataset contain audio samples with duration 1 second. 
For each audio sample, we add different kind of noises like dishwasher, running tap and exercise bike on half the volume of the speech sample. Then we use FFT to convert the audio samples to their frequency domain. The sampling rate of the speech files is 16kHz, so in the frequency domain it has 8000 components according to the Nyquist–Shannon sampling theorem~\parencite{Shannon1949}. A convolutional DNN is used to generate embedding representations of each audio wave in the frequency domain with size 32. In the case when the triplet architecture is used for the DNN's training, the semi-hard negatives mining produce the best results. Finally, the GTSRB dataset contains traffic sign images of variable sizes. In order to be able to use a single DNN to produce embedding representations for the image data, every image is either up-sampled by interpolation or down-sampled to 96x96x3. A convolutional DNN is used to generate embedding representations with size 128. In the triplet case, the training produced better results when mining for hard negatives is used.

%How to explain the DNN architectures

We first look at how well the distance metric learning methods cluster data of each class. A commonly used metric of the separation between classes is the \textit{Silhouette}~\parencite{ROUSSEEUW198753}. For each sample, we first compute the mean distance between $i$ and all other data points in the same cluster in the embedding space
$$a(i)=\frac{1}{|C_i|-1}\sum_{j\in C_i, i\neq j}d(i,j) \text{  .}$$
Then we compute the smallest mean distance from $i$ to all the data points in any other cluster
$$b(i)=\min_{k\notin i}\frac{1}{|C_k|}\sum_{j\in C_k}d(i,j) \text{  .}$$
The silhouette value is defined as
$$s(i)=\frac{b(i)-a(i)}{\max\{a(i),b(i)\}} \text{  .}$$
Each sample $i$ in the embedding space is assigned a silhouette value $-1\leq s(i)\leq 1$ depending on how close and how far it is to samples belonging to the same and different classes respectively. The closer $s(i)$ is to 1, the closer the sample is to samples of the same class and further from samples belonging to other classes. To compare the representations learned using the different methods as well as compute how much the clustering improves over the original inputs, we compute the mean silhouette over the training data and the validation data separately. In Table \ref{tab:silhouettes}, we see that the representations learned by either the siamese or the triplet network form well-defined clusters and are improved over the baseline clusters. On the other hand, the original inputs are not arranged in clusters.

\begin{table}[H]
\centering
\begin{tabular}{cc|c|c|}
\cline{3-4}
                                                           &                    & Training Silhouette & Validation Silhouette \\ \hline
\multicolumn{1}{|c|}{\multirow{3}{*}{SCITOS-G5}}           & Triplet Embeddings & 0.72                & 0.64                  \\ \cline{2-4} 
\multicolumn{1}{|c|}{}                                     & Siamese Embeddings & 0.94                & 0.8                   \\ \cline{2-4}
\multicolumn{1}{|c|}{}                                     & Baseline Embeddings & 0.23                & 0.23                  \\ \cline{2-4} 
\multicolumn{1}{|c|}{}                                     & Original Inputs           &   -0.03             &     -0.03          \\ \hline\hline
\multicolumn{1}{|c|}{\multirow{3}{*}{Speaker Recognition}} & Triplet Embeddings & 0.64                & 0.58                  \\ \cline{2-4} 
\multicolumn{1}{|c|}{}                                     & Siamese Embeddings & 0.8                 & 0.66                  \\ \cline{2-4}
\multicolumn{1}{|c|}{}                                     & Baseline Embeddings & 0.19                & 0.19                  \\ \cline{2-4} 
\multicolumn{1}{|c|}{}                                     & Original Inputs           & -0.03               & -0.03                 \\ \hline\hline
\multicolumn{1}{|c|}{\multirow{3}{*}{GTSRB}}               & Triplet Embeddings & 0.43                & 0.43                  \\ \cline{2-4} 
\multicolumn{1}{|c|}{}                                     & Siamese Embeddings & 0.75                & 0.72                  \\ \cline{2-4} 
\multicolumn{1}{|c|}{}                                     & Baseline Embeddings & 0.23                & 0.24                  \\ \cline{2-4}
\multicolumn{1}{|c|}{}                                     & Original Inputs           & -0.22               & -0.23                 \\ \hline
\end{tabular}
\caption{Clustering comparison using the silhouette coefficient}
\label{tab:silhouettes}
\end{table}

\subsection{Selecting the Significance Level}
\begin{figure}[H]
\centering
\includegraphics[width=0.6\linewidth]{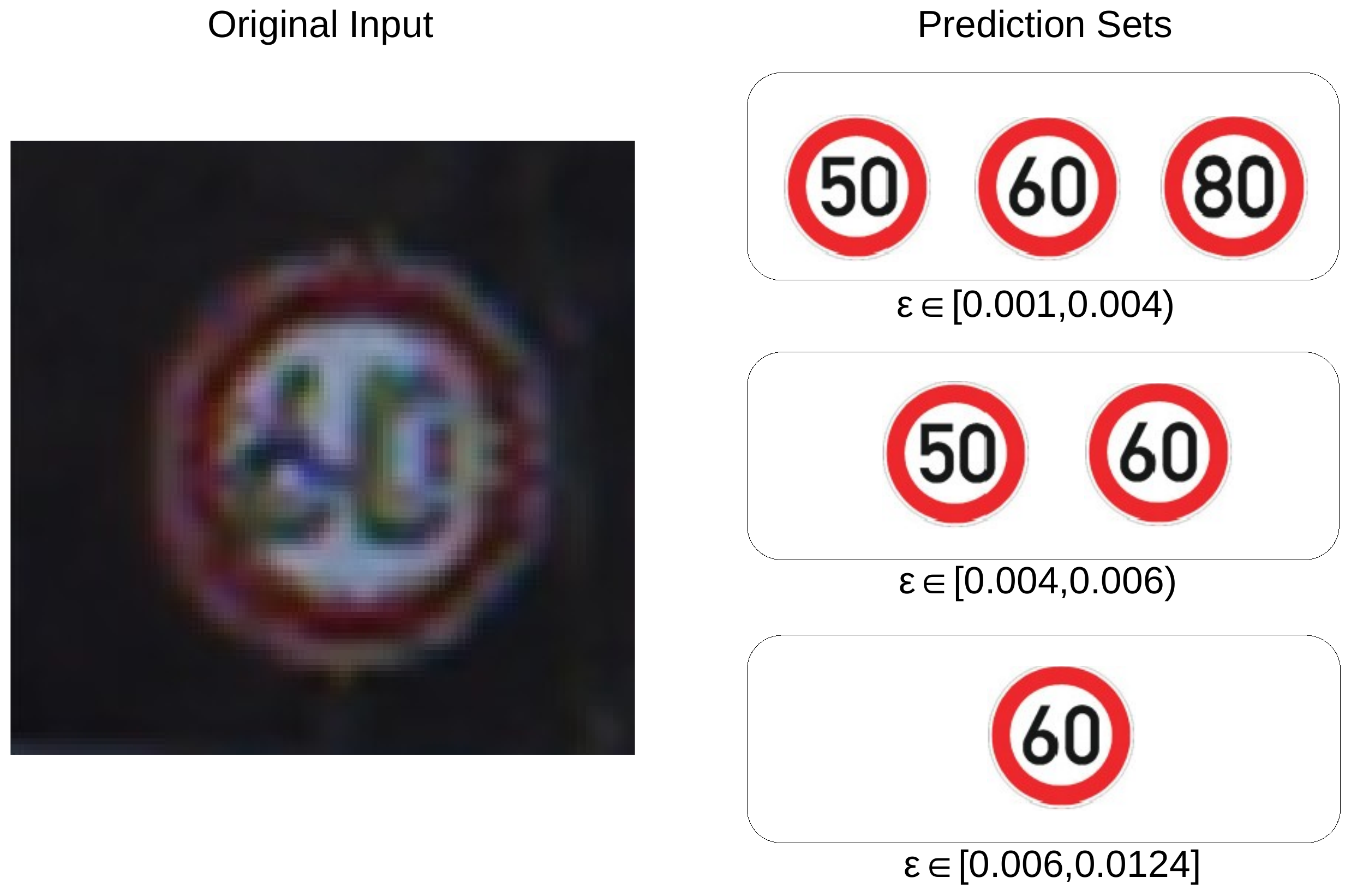}
\caption{Illustrative example}
\label{fig:sign_example}
\end{figure}

First, we illustrate the assurance monitoring algorithm with a test example from the GSTRB dataset. The left side of the Figure~\ref{fig:sign_example} shows the image of a 60km/h speed limit sign. Using nearest centroid as the NC function and the siamese network, Algorithm~\ref{alg:online} can be used to generate sets of possible predicted labels. In the following, we vary the significance level $ \epsilon $ and we report the set predictions. 
When $\epsilon \in [0.001,0.004)$, the possible labels are 'Speed limit 50km/h', 'Speed limit 60km/h', 'Speed limit 80km/h'; when $\epsilon \in [0.004,0.006)$, the possible labels are 'Speed limit 50km/h', 'Speed limit 60km/h'; and finally when $\epsilon\in[0.006,0.0124]$, the algorithm produces a single prediction 'Speed limit 60km/h' which is obviously correct.

\begin{figure}[H]
\centering
\includegraphics[width=1\linewidth]{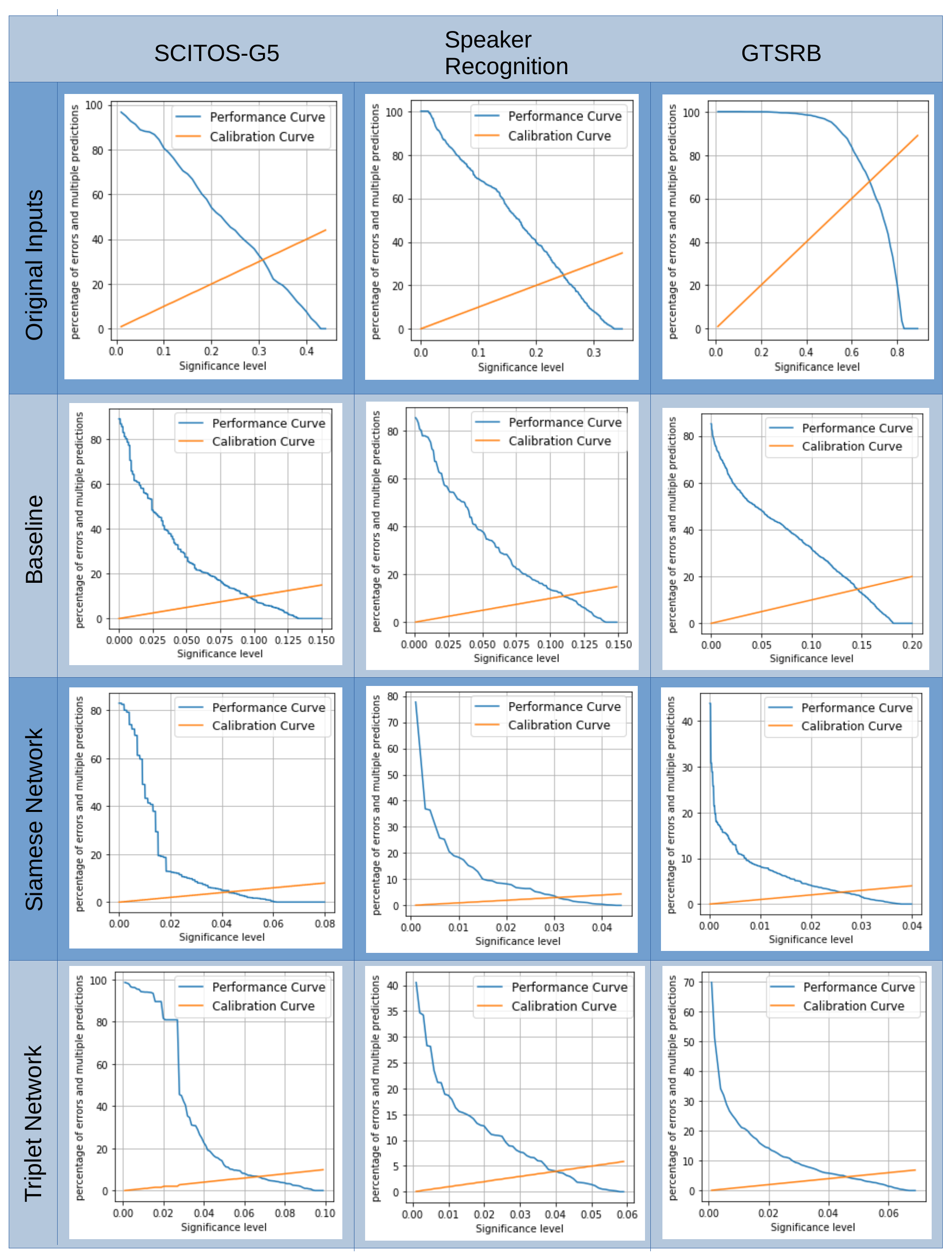}
\caption{Performance and calibration curves formed using the validation data from the different datasets using the nearest centroid NC function}
\label{fig:epsilon_estimation}
\end{figure}

For monitoring of CPS, one can either choose $\epsilon$ to be small enough given the system requirements or compute $ \epsilon $ to minimize the number of multiple predictions. Since the number of multiple predictions decreases when $\epsilon$ increases, we can select $\epsilon$ as the smallest value that eliminates multiple predictions for a calibration/validation set. This can be seen in Figure~\ref{fig:epsilon_estimation} where for each dataset, the optimal $\epsilon$ is selected as the significance level value where the performance curve goes to 0. The nearest centroid NC function is used for the plots in this figure.

Table~\ref{tab:epsilon_performance} shows the results for the different datasets and the various NC functions. First using the calibration/validation dataset, we select $\epsilon$ to eliminate sets of multiple predictions and we report the errors in the predictions for the testing dataset. The algorithm successfully did not generate any set with multiple predictions for the testing datasets for any of the NC functions other than the 1-NN when it was used in the SCITOS-G5 dataset with representations computed with the triplet network. In this particular case there was no $\epsilon$ that could eliminate the prediction sets with multiple classes and even when $\epsilon=1$, 38.6\% of the test inputs produced prediction sets with multiple classes. The error-rates are well-calibrated and bounded by the computed or the chosen significance level. One way to compare the different NCMs is by looking at the significance level that is required for ICP to make single predictions. The use of embedding representations could always produce single predictions using significance levels much lower than when the original inputs are used. The significance of the distance metric learning techniques is apparent in the case of the nearest centroid NCM on all the datasets. This is an appealing NCM for its simplicity and the reduced memory requirements. When used as part of the baseline the performance was not as good as the more expensive NCMs. However, leveraging the better clustering that distance metric learning methods achieve, the nearest centroid NCM performs as well or better than the rest of the NCMs on making predictions with low significance level while retaining the computational efficiency. We also evaluate how well the different approaches bound the error-rate for two different values of the significance level. The errors are bounded in most cases no matter if embedding representations are used or not. The percentage of set predictions on the test data that have multiple candidate classes tend to increase the lower the chosen $\epsilon$ is compared to the estimated optimal $\epsilon$.

\begin{table}[H]
\centering
\begin{tabular}{ccc|c|c|c|c|c|c|}
\cline{4-9}
                                                           &                                                &                  & \multicolumn{2}{c|}{Estimate $\epsilon$} & \multicolumn{2}{c|}{$\epsilon$=0.01} & \multicolumn{2}{c|}{$\epsilon$=0.02} \\ \hline
\multicolumn{1}{|c|}{Dataset}                              & \multicolumn{1}{c|}{Architecture}              & NC Function      & $\epsilon$            & Errors           & Errors                 & Multiples                & Errors                 & Multiples                \\ 
% \hhline{|=|=|=|=|=|=|=|=|=|}
\hline\hline
\multicolumn{1}{|c|}{\multirow{9}{*}{SCITOS-G5}}           & \multicolumn{1}{c|}{\multirow{3}{*}{Triplet}}  & $k$-NN           &          0.087                     &        9.2\%     &      0\%               &        100\%             &         0\%               &     100\%                \\ \cline{3-9} 
\multicolumn{1}{|c|}{}                                     & \multicolumn{1}{c|}{}                          & 1-NN             &          1.0                       &       61.4\%     &    0.4\%               &       88.3\%             &        1\%             &       71.1\%             \\ \cline{3-9} 
\multicolumn{1}{|c|}{}                                     & \multicolumn{1}{c|}{}                          & Nearest Centroid &           0.095                    &       8.4\%      &    0\%                 &       96.2\%             &        0.5\%           &       84.6\%            \\ \cline{2-9} 
\multicolumn{1}{|c|}{}                                     & \multicolumn{1}{c|}{\multirow{3}{*}{Siamese}}  & $k$-NN           &           0.066                    &       6.6\%      &    0\%                 &       100\%              &        0\%             &       100\%              \\ \cline{3-9} 
\multicolumn{1}{|c|}{}                                     & \multicolumn{1}{c|}{}                          & 1-NN             &           0.078                    &       8.2\%      &    0.4\%               &       71.8\%             &        2.2\%           &       21.4\%             \\ \cline{3-9} 
\multicolumn{1}{|c|}{}                                     & \multicolumn{1}{c|}{}                          & Nearest Centroid &           0.062                    &       7.1\%      &    0.2\%               &       45.8\%             &        1.5\%           &       13.4\%             \\ \cline{2-9} 
\multicolumn{1}{|c|}{}                                     & \multicolumn{1}{c|}{\multirow{3}{*}{Baseline}}  & $k$-NN           &           0.093                    &       7.9\%      &    0\%                 &       100\%              &      0.7\%             &      36.3\%              \\ \cline{3-9} 
\multicolumn{1}{|c|}{}                                     & \multicolumn{1}{c|}{}                          & 1-NN             &           0.074                    &       7.5\%      &    0.9\%               &       35.7\%             &        1.3\%           &       27.8\%             \\ \cline{3-9} 
\multicolumn{1}{|c|}{}                                     & \multicolumn{1}{c|}{}                          & Nearest Centroid &           0.133                    &     16.1\%      &    0.7\%               &       67.9\%             &        1.6\%           &       55.9\%             \\ \cline{2-9} 
\multicolumn{1}{|c|}{}                                     & \multicolumn{1}{c|}{\multirow{3}{*}{\parbox{1.5cm}{Original\\ Inputs}}} & $k$-NN           &           0.198                    &       22.3\%    &       0.7\%            &             72.1\%       &         1.6\%          &        58.4\%            \\ \cline{3-9} 
\multicolumn{1}{|c|}{}                                     & \multicolumn{1}{c|}{}                          & 1-NN             &            0.122                   &      12.6\%      &    1\%                 &           57.9\%         &            3.7\%       &       37.5\%             \\ \cline{3-9} 
\multicolumn{1}{|c|}{}                                     & \multicolumn{1}{c|}{}                          & Nearest Centroid &           0.428                    &       43.5\%     &      0.5\%             &          96.9\%          &          0.7\%         &         95.8\%           \\ \hline\hline
\multicolumn{1}{|p{1.5cm}|}{\multirow{9}{*}{\parbox{1.5cm}{Speaker\\ Recognition}}} & \multicolumn{1}{c|}{\multirow{3}{*}{Triplet}}  & $k$-NN                       &       0.058      &      5.2\%             &          1.5\%           &        16.6\%          &        2.5\%           &       10\%               \\ \cline{3-9} 
\multicolumn{1}{|c|}{}                                     & \multicolumn{1}{c|}{}                          & 1-NN             &           0.063                    &       5.9\%      &    2.1\%               &       22.6\%             &        2.8\%           &       14.1\%             \\ \cline{3-9} 
\multicolumn{1}{|c|}{}                                     & \multicolumn{1}{c|}{}                          & Nearest Centroid &           0.058                    &       6.3\%      &    1.9\%               &       20.1\%             &        2.5\%           &       14\%               \\ \cline{2-9} 
\multicolumn{1}{|c|}{}                                     & \multicolumn{1}{c|}{\multirow{3}{*}{Siamese}}  & $k$-NN           &           0.041                    &       3.6\%      &    0\%                 &       100\%              &        2.3\%           &       7.3\%              \\ \cline{3-9} 
\multicolumn{1}{|c|}{}                                     & \multicolumn{1}{c|}{}                          & 1-NN             &           0.045                    &       4.5\%      &    0.9\%               &       16.1\%             &        2\%             &       7.6\%              \\ \cline{3-9} 
\multicolumn{1}{|c|}{}                                     & \multicolumn{1}{c|}{}                          & Nearest Centroid &           0.043                    &       4\%        &    1.5\%               &       14.8\%             &        1.9\%           &       7.1\%              \\ \cline{2-9} 
\multicolumn{1}{|c|}{}                                     & \multicolumn{1}{c|}{\multirow{3}{*}{Baseline}}  & $k$-NN           &           0.03                     &       3.1\%      &  0.9\%                 &       6.7\%              &      1.6\%             &      2.5\%              \\ \cline{3-9} 
\multicolumn{1}{|c|}{}                                     & \multicolumn{1}{c|}{}                          & 1-NN             &           0.033                    &       3.6\%      &    0.9\%               &        6.9\%             &        2.3\%           &        2.9\%             \\ \cline{3-9} 
\multicolumn{1}{|c|}{}                                     & \multicolumn{1}{c|}{}                          & Nearest Centroid &           0.141                    &      14.6\%      &    0.5\%               &       78.2\%             &        2.1\%           &       61.9\%             \\ \cline{2-9} 
\multicolumn{1}{|c|}{}                                     & \multicolumn{1}{c|}{\multirow{3}{*}{\parbox{1.5cm}{Original\\ Inputs}}} & $k$-NN           &           0.295                    &       29.6\%     &    0\%                 &        100\%             &        0\%             &      100\%               \\ \cline{3-9} 
\multicolumn{1}{|c|}{}                                     & \multicolumn{1}{c|}{}                          & 1-NN             &           0.231                    &       22.9\%     &    0.7\%               &        84.82\%           &        1.3\%           &      76.7\%              \\ \cline{3-9} 
\multicolumn{1}{|c|}{}                                     & \multicolumn{1}{c|}{}                          & Nearest Centroid &           0.336                    &       33.3       &    0.7\%               &       100\%              &        1.6\%           &       98.3\%             \\ \hline\hline
\multicolumn{1}{|c|}{\multirow{9}{*}{GTSRB}}               & \multicolumn{1}{c|}{\multirow{3}{*}{Triplet}}  & $k$-NN           & 0.04                               & 4.8\%            & 1.5\%                  & 9.3\%                    & 2.6\%                  & 4.6\%                    \\ \cline{3-9} 
\multicolumn{1}{|c|}{}                                     & \multicolumn{1}{c|}{}                          & 1-NN             &  0.031                             &  3.8\%           &  1.4\%                 &  8.2\%                   &   2.7\%                &   2.9\%                  \\ \cline{3-9} 
\multicolumn{1}{|c|}{}                                     & \multicolumn{1}{c|}{}                          & Nearest Centroid & 0.067                              & 6.3\%            & 1.5\%                  & 22.6\%                   & 2.5\%                  & 13.9\%                   \\ \cline{2-9} 
\multicolumn{1}{|c|}{}                                     & \multicolumn{1}{c|}{\multirow{3}{*}{Siamese}}  & $k$-NN           &    0.031                           &  3.1\%           &   0.9\%                &   7.4\%                  &     1.8\%             &    3.4\%                 \\ \cline{3-9} 
\multicolumn{1}{|c|}{}                                     & \multicolumn{1}{c|}{}                          & 1-NN             &    0.035                           &   3.3\%          &    0.8\%               &    9.5\%                 &      1.5\%             &   4.8\%                  \\ \cline{3-9} 
\multicolumn{1}{|c|}{}                                     & \multicolumn{1}{c|}{}                          & Nearest Centroid &    0.038                           &     3.8\%        &    1.2\%               &    8.7\%                &        2.5\%           &     4.1\%                \\ \cline{2-9} 
\multicolumn{1}{|c|}{}                                     & \multicolumn{1}{c|}{\multirow{3}{*}{Baseline}}  & $k$-NN           &           0.011                    &       1.1\%      &    1\%                 &       0.1\%              &        2\%             &       0\%              \\ \cline{3-9} 
\multicolumn{1}{|c|}{}                                     & \multicolumn{1}{c|}{}                          & 1-NN             &           0.003                    &       0.3\%      &      1\%               &          0\%             &        2.1\%           &          0\%             \\ \cline{3-9} 
\multicolumn{1}{|c|}{}                                     & \multicolumn{1}{c|}{}                          & Nearest Centroid &           0.182                    &      19.4\%      &    0.9\%               &       71.8\%             &        1.9\%           &       62.8\%             \\ \cline{2-9} 
\multicolumn{1}{|c|}{}                                     & \multicolumn{1}{c|}{\multirow{3}{*}{\parbox{1.5cm}{Original\\ Inputs}}} & $k$-NN           &    0.731                           &     71.7\%       &    1.2\%               &    98.5\%                &       1.2\%            &    98.5\%                \\ \cline{3-9} 
\multicolumn{1}{|c|}{}                                     & \multicolumn{1}{c|}{}                          & 1-NN             &     ---                            &    ---           &     ---                &         ---              &       ---              &       ---                \\ \cline{3-9} 
\multicolumn{1}{|c|}{}                                     & \multicolumn{1}{c|}{}                          & Nearest Centroid &     0.824                          &      82.7\%      &    0.9\%               &      100\%               &      2\%               &      100\%               \\ \hline
\end{tabular}
\caption{ICP performance for the different configurations}
\label{tab:epsilon_performance}
\end{table}

\subsection{Computational Efficiency}
\label{sec:computation_efficiency}
In order to evaluate if the approach can be used for real-time monitoring of CPS, we measure the execution times and the memory requirements. Different NC functions lead to different execution times and memory requirements. We compare the average execution time over the testing datasets required for generating a prediction set after the model receives a new test input in Table \ref{tab:time_memory}. The $1$-NN NC function on the input space of the GTSRB dataset has excessive memory requirements. Below we present the computational requirements for each NC function and explain the higher requirements of the $1$-NN function in more detail.

\begin{table}[H]
\centering
\begin{tabular}{|c|c|c|c|c|}
\hline
Dataset                              & Architecture              & NC Function      & Execution Time & Memory   \\ \hline\hline
\multirow{9}{*}{SCITOS-G5}           & \multirow{3}{*}{Triplet}  & k-NN             &  0.2ms         & 700.6 kB \\ \cline{3-5} 
                                     &                           & 1-NN             &  1.8ms         & 2 MB     \\ \cline{3-5} 
                                     &                           & Nearest Centroid &  56$\mu$s      & 324.8 kB \\ \cline{2-5} 
                                     & \multirow{3}{*}{Siamese}  & k-NN             &  0.2ms         & 700.6 kB \\ \cline{3-5} 
                                     &                           & 1-NN             &  1.6ms         & 2 MB     \\ \cline{3-5} 
                                     &                           & Nearest Centroid &  56$\mu$s      & 324.8 kB \\ \cline{2-5}
                                     & \multirow{3}{*}{Baseline} & k-NN             & 0.2ms          & 700.6 kB  \\ \cline{3-5} 
                                     &                           & 1-NN             & 1.6ms          &  2 MB    \\ \cline{3-5} 
                                     &                           & Nearest Centroid & 58$\mu$s       & 324.8 kB \\ \cline{2-5}
                                     & \multirow{3}{*}{\parbox{1.5cm}{Original\\ Inputs}} & k-NN             &  0.3ms         & 2.4 MB   \\ \cline{3-5} 
                                     &                           & 1-NN             &  1.9ms         & 4.1 MB   \\ \cline{3-5}
                                     &                           & Nearest Centroid &  59$\mu$s      & 763.1 kB \\ \hline\hline
\multirow{9}{*}{Speaker Recognition} & \multirow{3}{*}{Triplet}  & k-NN             &  0.2ms         & 15.3 MB  \\ \cline{3-5} 
                                     &                           & 1-NN             &  2.1ms         & 24.6 MB \\ \cline{3-5} 
                                     &                           & Nearest Centroid &  74$\mu$s      & 13.1 MB  \\ \cline{2-5} 
                                     & \multirow{3}{*}{Siamese}  & k-NN             & 0.2ms          & 15.3 MB  \\ \cline{3-5} 
                                     &                           & 1-NN             & 2ms            & 24.6 MB  \\ \cline{3-5} 
                                     &                           & Nearest Centroid & 0.1ms          & 13.1 MB  \\ \cline{2-5}
                                     & \multirow{3}{*}{Baseline} & k-NN             & 0.3ms          & 15.3 MB  \\ \cline{3-5} 
                                     &                           & 1-NN             & 2.3ms          &  24.6 MB  \\ \cline{3-5} 
                                     &                           & Nearest Centroid & 71$\mu$s       & 13.1 MB  \\ \cline{2-5}
                                     & \multirow{3}{*}{\parbox{1.5cm}{Original\\ Inputs}} & k-NN             & 53ms           & 723.9 MB \\ \cline{3-5} 
                                     &                           & 1-NN             & 274ms          & 2.3 GB   \\ \cline{3-5} 
                                     &                           & Nearest Centroid & 0.1ms          & 378.7 MB \\ \hline\hline
\multirow{9}{*}{GTSRB}               & \multirow{3}{*}{Triplet}  & k-NN             & 0.6ms          & 70.1 MB  \\ \cline{3-5} 
                                     &                           & 1-NN             & 24.6ms         & 1.4 GB   \\ \cline{3-5} 
                                     &                           & Nearest Centroid & 0.7ms          & 38.4 MB  \\ \cline{2-5} 
                                     & \multirow{3}{*}{Siamese}  & k-NN             & 0.5ms          & 70.1 MB  \\ \cline{3-5} 
                                     &                           & 1-NN             & 21.8ms         & 1.4 GB   \\ \cline{3-5} 
                                     &                           & Nearest Centroid & 0.6ms          & 38.4 MB  \\ \cline{2-5} 
                                     & \multirow{3}{*}{Baseline} & k-NN             & 0.6ms          & 70.1 MB   \\ \cline{3-5} 
                                     &                           & 1-NN             & 24.4ms         &  1.4 GB  \\ \cline{3-5} 
                                     &                           & Nearest Centroid & 0.7ms          & 38.4 MB \\ \cline{2-5}
                                     & \multirow{3}{*}{\parbox{1.5cm}{Original\\ Inputs}} & k-NN             & 654ms          & 8.9 GB   \\ \cline{3-5} 
                                     &                           & 1-NN             & ---            &  ---     \\ \cline{3-5} 
                                     &                           & Nearest Centroid & 4.8ms          & 2.1 GB   \\ \hline
\end{tabular}
\caption{Execution times and memory requirements}
\label{tab:time_memory}
\end{table}

Table~\ref{tab:time_memory} reports the average execution time for each test input and the required memory space using different NC functions. 
Datasets with high-dimensional inputs are challenging for applying ICP in real-time and the results demonstrate the impact of the embedding representations use on the execution times. All the NC functions require storing the the calibration NC scores which are used for computing the test p-values online. The DNN weights need to be stored when embedded representations need to be calculated for every new test input. Furthermore each NCM has a different memory overhead. In the $k$-NN case, the encodings of the training data are stored in a $k-d$ tree~\parencite{Bentley:1975:MBS:361002.361007} that is used to compute efficiently the $k$ nearest neighbors. This data structure is used both for the $k$-NN and $1$-NN NC functions. In the $1$-NN case, it is required to find the nearest neighbor in the training data for each possible class which is computationally expensive resulting in larger execution time. The nearest centroid NC function requires storing only the centroids for each class and the additional memory required is minimal.

In conclusion, the evaluation results demonstrate that monitoring based on ICP has well-calibrated error rates in all configurations. Further the use of embedding representations reduces the computational requirements and can lead to decisions with improved significance level. Using distance metric learning methods the training data form well-defined clusters that is essential in the case of the nearest centroid NCM. This improvement makes it a good NCM option for all of the used datasets as it performs as well as the other NCMs but with significantly less computational requirements.

\section{Concluding Remarks}
\label{sec:conclusion}
Cyber-physical systems (CPS) incorporate machine learning components such as DNNs for performing various tasks such as perception of the environment.
When used for safety-critical applications they need to satisfy specific requirements that are defined taking into account the acceptable risk and its cost for incorrect decisions.
Although DNNs offer advanced capabilities on the decision making process, they cannot provide guarantees on the estimated error-rate. To achieve this they must be complemented by engineering methods and 
practices that allow effective integration in CPS where an accurate estimate of confidence is needed. 

The paper considers the problem of complementing the prediction of DNNs
 with a well-calibrated confidence. For classification tasks, the inductive conformal prediction framework allows selecting the significance level according to the requirements of each application. This is a parameter that defines the acceptable error-rate and is a trade-off between errors and alarms. We presented computationally efficient algorithms based on representations learned by underlying DNN models that make possible for ICP to be used for real-time monitoring.
 The proposed approach was evaluated on three different benchmarks of increasing complexity from a mobile robot with ultrasound sensors, to speaker recognition and traffic sign recognition.
The evaluation results demonstrate that monitoring based on the inductive conformal prediction framework using embedding representations instead of the original inputs has well-calibrated error-rates and can minimize the number of alarms when a confident decision cannot be made. When appropriate embedding representations are computed using distance metric learning methods input data that belong to the same class form well-defined clusters. This property is very important when the similarity of a test input to the test data is estimated.
That way the training set can be efficiently represented by the centroids of each class which reduces the computational requirements without any loss in performance when compared to the more computationally expensive approaches. 

During the experiments we identified a number of challenges that can lead to poor performance of the proposed method. First, when the datasets are imbalanced both the siamese and the triplet architectures may not learn embedding representations that cluster the under-represented classes well. This affects the efficiency of the NC functions. Second, the training of the triplet networks require mining of training data that will form triplets that lead to large gradients for minimizing the triplet loss function. There is ongoing research for mining algorithms for faster training. 
One open question for future research is how to utilize all the candidate decisions in the prediction set to deal with the cases when a confident decision cannot be made that will satisfy the significance-level requirements.

\vspace{1em}

\noindent 
\textbf{Dimitrios Boursinos} is currently pursuing his Ph.D. degree in Electrical Engineering at Vanderbilt University. He is also a research assistant in the Institute for Software Integrated Systems (ISIS). His research interests focus on machine learning and cyber-physical systems with emphasis on learning-enabled systems.

\vspace{1em}
\noindent 
\textbf{Xenofon Koutsoukos} is a Professor of Computer Science, Computer Engineering, and Electrical Engineering and Chair of the Department of Electrical Engineering and Computer Science at Vanderbilt. He is also a Senior Research Scientist in the Institute for Software Integrated Systems (ISIS). His research work is in the area of cyber-physical systems with emphasis on learning-enabled systems, security and resilience, control and decision making, diagnosis and fault tolerance, formal methods, and adaptive resource management. He is a Fellow of the IEEE for his contributions to the design of resilient cyber-physical systems.

\clearpage
% \bibliographystyle{apacite}
% \bibliography{main}
\printbibliography
\end{document}